\documentclass[twoside,11pt]{article}

\usepackage{blindtext}

\usepackage[square,numbers]{natbib}
\usepackage{dirtytalk}          
\usepackage{xcolor}             
\usepackage{colortbl}           

\usepackage{amssymb}        
\usepackage{mathtools}      
\usepackage{amsthm}         

\usepackage{graphicx}       
\usepackage{tikz}
\usetikzlibrary{shapes,arrows}

\usepackage{algorithm}
\usepackage[noend]{algpseudocode}

\usepackage{url}

\usepackage{appendix}

\usepackage{multirow} 

\newcommand{\indic}[1]{\mathbf{1}_{#1}}                         
\newcommand{\defeq}{\vcentcolon =}                              
\newcommand{\expecunder}[2]{\mathbb{E}_{#2}\left[#1\right]}     
\renewcommand{\epsilon}{\varepsilon}
\newcommand*{\Maximize}{\mathrm{Maximize}}     
\newcommand*{\Minimize}{\mathrm{Minimize}}     
\newcommand{\expec}[1]{\mathbb{E}\left[#1\right]}               
\newcommand{\proba}[1]{\mathbb{P}\left (#1\right )}             
\newcommand{\logs}[1]{\log{\left(#1\right)}}         
\newcommand{\ceil}[1]{\left\lceil#1\right\rceil}           
\newcommand{\Reals}{\mathbb{R}}                                 
\newcommand{\cvproba}{\overset{\mathbb{P}}{\longrightarrow}}
\newcommand{\probaunder}[2]{\mathbb{P}_{#2}\left (#1\right )}   

\newcommand{\doc}{z}                                            
\newcommand{\texts}{\mathcal{T}}                                
\newcommand{\word}{w}                                           
\newcommand{\dictionary}{\mathcal{D}}                           
\newcommand{\localdic}{\mathcal{D}}                             
\newcommand{\drop}{d}                                           
\newcommand{\candrop}{\Delta}                                           
\newcommand{\candidate}{c}                                      
\newcommand{\candidates}{\mathcal{C}}                           
\newcommand{\size}[1]{\left\lvert#1\right\rvert}                
\newcommand{\Empdrop}{\widehat{\candrop}}                  
\newcommand{\length}{\ell}                                      
\newcommand{\idf}{v}                                            
\newcommand{\Tfidf}{\varphi}
\newcommand{\tfidf}[1]{\Tfidf(#1)}                              
\newcommand{\mult}{m}                                           
\newcommand{\Mult}{M}                                           
\newcommand{\card}[1]{\left\lvert#1\right\rvert}                
\newcommand{\explanation}{e}                
\newcommand{\model}{f}                                      

\usepackage{nicematrix}
\usepackage{booktabs}

\usepackage{tikz}

\usepackage{jmlr2e}

\usepackage{lastpage}

\firstpageno{1}

\begin{document}

\title{Faithful and Robust Local Interpretability \\for Textual Predictions}

\author{\name Gianluigi Lopardo \email glopardo@unice.fr \\
       \addr Université Côte d'Azur, Inria, CNRS \\
       Laboratoire Jean-Alexandre Dieudonné \\
       Nice, France
       \AND
       \name Frédéric Precioso \email frederic.precioso@inria.fr \\
       \addr Université Côte d'Azur, Inria, CNRS \\
       Laboratoire d’Informatique, Signaux et Systèmes de Sophia Antipolis \\
       Nice, France
       \AND
       \name Damien Garreau \email damien.garreau@uni-wuerzburg.de \\
       \addr
        Julius-Maximilians Universit\"at \\
        Institute of Computer Science \\
        W\"urzburg, Germany
}

\maketitle

\begin{abstract}%
Interpretability is essential for machine learning models to be trusted and deployed in critical domains. However, existing methods for interpreting text models are often complex, lack mathematical foundations, and their performance is not guaranteed. 
In this paper, we propose FRED (Faithful and Robust Explainer for textual Documents), a novel method for interpreting predictions over text. 
FRED offers three key insights to explain a model prediction: (1) it identifies the minimal set of words in a document whose removal has the strongest influence on the prediction, (2) it assigns an importance score to each token, reflecting its influence on the model's output, and (3) it provides counterfactual explanations by generating examples similar to the original document, but leading to a different prediction. 
We establish the reliability of FRED through formal definitions and theoretical analyses on interpretable classifiers. Additionally, our empirical evaluation against state-of-the-art methods demonstrates the effectiveness of FRED in providing insights into text models. 
\end{abstract}

\begin{keywords}
    Explainable AI, Interpretability, Natural Language Processing, Text Classification
\end{keywords}

\section{Introduction}
\label{sec:intro}

Interpretability is essential for machine learning models to be trusted and deployed in critical and sensitive contexts, such as in medical or legal domains~\citep{carvalho2019machine}. 
Local and model-agnostic methods are particularly well-suited for this task because they can explain predictions made by any model for a specific instance without requiring any knowledge about the underlying model \citep{ribeiro2016should,lundberg2017unified,guidotti2018local,ribeiro2018anchors,montavon2019layer}. 
This makes them more versatile and applicable to a wider range of scenarios than other classes of methods, which typically intervene during model training \citep{ciravegna2021logic,rigotti2021attention} or otherwise need access to some model parameters \citep{selvaraju2017grad,mylonas2023attention}. 

\begin{figure}[t]
\begin{minipage}{0.5\textwidth}
    (a) minimal influential subset of tokens \\
    \texttt{Explaining class \say{positive}:} \\
    \texttt{The minimal subset of tokens that make the confidence drop by $50.0\%$ if perturbed is
    	\{\say{decent}, \say{great}\}} \\
\end{minipage}
\hfill
\begin{minipage}{0.4\textwidth}
    (b) per-token importance score \\
    \includegraphics[scale=0.6]{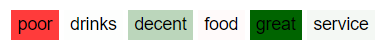}
\end{minipage}

(c) counterfactual explanations \\
\texttt{$k=3$ samples with minimal perturbation classified as \say{negative}:} \\
\texttt{\say{poor drinks decent food \color{orange}dirty \color{black} service}} \\
\texttt{\say{poor drinks decent food \color{orange}awful \color{black} service}} \\
\texttt{\say{poor drinks \color{orange}bad \color{black} food great service}}
\caption{\label{fig:fred-exp}FRED explaining the prediction of a sentiment analysis model for the restaurant review “{poor drinks, decent food, great service}”, classified as “positive”. 
The average confidence over the sample is $0.556$. 
\textbf{(a)} FRED identifies the minimal subset of tokens that, if removed, make the prediction drop by a specified threshold $\epsilon(=0.5$). 
\textbf{(b)} Saliency map of token importance score: dark green (resp., red) means high positive (resp., negative) influence. 
\textbf{(c)} Samples close to the example, but classified as "negative". 
Perturbations with respect to the example are in orange. 
}
\end{figure}

Nonetheless, several interpretability methods are afflicted by an absence of theoretical basis. 
In particular, it is often unclear how these methods perform on simple, already interpretable models \citep{garreau2020explaining}. 
Additionally, each explainer is characterized by a number of internal mechanisms (\emph{e.g.}, sampling, local approximations, measures of importance) that can have a very different impact on the final explanations \citep{covert_et_al_2021}. 
These mechanisms are often ignored or little studied, making the explainers in themselves just as mysterious as the prediction to be explained. 

Thus, instead of providing clarity, using an explainer that is poorly understood on a complex model can lead to misinterpretations of the model's behavior~\citep{lipton2018mythos}. 
For example, consider the case of an automated loan application. 
The bank operator can use an explainer to understand which features impacted the most on a decision, and validate or reject it. 
Some explainers define as more important features those closer to the decision boundary, \emph{e.g.}, those with values just below or above certain thresholds. 
Others, however, tend to highlight features with more extreme values, which thus strongly impact a decision in a different sense. 
Since there is no mathematical or legal agreement on what a good explanation should be, it is important to have a clear idea of the method used in order to draw the right conclusions. 

We believe that text models are a particularly understudied area of machine learning interpretability. 
Despite their increasing complexity and prevalence, interpretability studies have not kept pace with the advances that have resulted from transformers and large language models, consisting nowadays in billions or trillions of parameters \citep{devlin2019bert, brown_et_al_2020, chowdhery2023palm, touvron2023llama, minaee2024large}. 

In this paper, we introduce FRED (Faithful and Robust Explainer for text Documents): a novel interpretability framework for text classification and regression tasks. 
FRED offers three key insights to explain a model prediction: (1) it identifies the minimal set of words in a document whose removal has the strongest influence on the prediction, (2) it assigns an importance score to each token, reflecting its influence on the model's output, and (3) it provides counterfactual explanations by generating examples similar to the original document, but leading to a different prediction. 
A simple illustration of FRED applied to a sentiment analysis task is shown in Figure~\ref{fig:fred-exp}. 

\paragraph{Organization of the paper.}
In the rest of this paper, we describe FRED in detail, validate its reliability, compare it to other interpretability methods, and discuss its advantages, as follows. 
First, in Section~\ref{sec:related} we present some related literature, to position our work. 
We then delve into the description of FRED in Section~\ref{sec:FRED}, by providing a formal definition of FRED's mechanics, elucidating the essence of its explanations. 
In particular, we illustrate a novel sampling scheme that leverage tokens' \emph{part-of-speech} tag. 
We then conduct a rigorous theoretical analysis of FRED's behavior on interpretable classifiers, ensuring that it aligns with expectations on simpler models in Section~\ref{sec:analysis}. 
In Section~\ref{sec:experiments}, we empirically evaluate FRED against well-established explainers on a variety of models, including state-of-the-art models, demonstrating its effectiveness, especially on more complex models and larger documents.
We draw our conclusion in Section~\ref{sec:conclusion}. 
The empirical results highlight an interesting trajectory: FRED performs better on more modern models, and larger documents, making it more suitable for realistic cases. 
We prove all our theoretical claims in Appendix~\ref{sec:proofs} and support them with numerical experiments, detailed in Appendix~\ref{appendix-sec:experiments}. 
The code for FRED and the experiments is available at \url{https://github.com/gianluigilopardo/fred}.

\subsection{Related work}
\label{sec:related}
Within the field of machine learning interpretability \citep{guidotti2018local,adadi2018peeking,linardatos2021explainable}, our work falls under the category of \emph{local post-hoc} methods. 
These methods are termed \emph{local} because they explain the prediction for a single data point, as opposed to explaining the model's overall behavior. 
They are \emph{post-hoc}  as they are applied to an already trained model, without needing access to its internal parameters. 
Conversely, other approaches focus on building inherently interpretable models or leveraging interpretable component \citep{ciravegna2021logic, selvaraju2017grad, 
lopardo2022smace}. 

Model-agnostic explainers work on any black-box model, requiring only repeated queries. 
Most, like LORE \citep{guidotti2018local} and LIME \citep{ribeiro2016should}, approximate the model locally for a specific instance, employing respectively a decision tree and a linear model as a local surrogate. 
Anchors \citep{ribeiro2018anchors} differ, extracting provable rules that guarantee the model's prediction. 

Local explainers leverage sampling, but risk generating out-of-distribution (OOD) data \citep{hase2021out}, thus causing inaccurate explanations and being vulnerable to adversarial attacks \citep{slack2020fooling}. 
Recent work addresses sampling issues. 
For instance, \citet{delaunay2020improving} improves Anchor sampling for tabular data, while \citet{amoukou2021consistent} extends Minimal Sufficient Rules (similar to Anchors) to regression models, handling continuous features without discretization.

While LIME and SHAP \citep{lundberg2017unified} explain predictions by feature importance, Anchors identifies a compact set of features guaranteed to produce the same prediction with high probability (details in \citet{lopardo2022sea}). 

Studies show that users prefer rule-based explanations \citep{lim2009and, stumpf2007toward}, such as
hierarchical decision lists \citep{wang2015falling}, revealing global behavior, and \citep{lakkaraju2016interpretable} balance accuracy and interpretability with smaller, disjoint rules. 

While most interpretability methods target structured data, a limited subset focuses on text \citep{danilevsky2020survey}. 
LIME, SHAP, and Anchors address text classification. 
LIME and SHAP assign importance scores to tokens, while Anchors target the most significant token set for the prediction. 
Our method, FRED, accomplishes both. 

Some works highlight the potential of counterfactual explanations. 
These explanations delve into what changes to the input text would cause a different model prediction, offering valuable insights into model behavior. 
\citet{wachter2017counterfactual, pawelczyk2021carla, pawelczyk2022exploring} explore this concept. 
Their work particularly emphasizes the importance of generating not only accurate but also plausible counterfactuals, which builds trust in the model's decision-making process. 

FRED leverages the \emph{explaining by removing} strategy, where removing features reveals their influence on predictions. 
While established for tabular data and images \citep{covert_et_al_2021}, it is underexplored for text. 
FRED isolates token sets within a document and measures the confidence drop upon removal. 
This quantifies the impact of each token. 
FRED pinpoints a concise subset of words crucial for the prediction by identifying those that lead to a substantial confidence drop when removed. 
Additionally, it assigns an importance value to each token that reflects its influence on the model's output. 
Finally, FRED offers counterfactual explanations by generating examples similar to the original document, but leading to a different prediction. 
This allows users to see which slight changes to the text can alter the model's decision-making process. 

The field of interpretable machine learning often prioritizes practical application over formal guarantees, leading to explanations that may not be reliable \citep{marques2022delivering}. 
To address this, \citet{la2021guaranteed} propose a method for generating robust explanations in text models, focusing on minimal word subsets sufficient for prediction and resistant to minor input changes. 
Inspired by this line of work \citep{garreau2020explaining}, and its adaptation to text data for LIME and Anchors \citep{mardaoui_garreau_2021, lopardo2022sea}, we perform a theoretical analysis to ensure FRED behave as expected on well-understood models like linear models and shortcut detection. 
This check is crucial to guarantee explanations reflect the model's true inner workings.

\section{FRED}
\label{sec:FRED}
This section introduces FRED, our novel explainer designed to provide faithful and robust explanations for text classification and regression tasks. 
FRED leverages a perturbation-based approach, analyzing the model's behavior on slightly altered versions of the original text. 
When presented with an example to explain, FRED first generates a perturbed sample (as detailed in Section~\ref{sec:sampling}). 
Then, FRED explains model predictions through three key functionalities: 
\begin{enumerate}
    \item It identifies the \textbf{minimal influential subset} of tokens within the example. 
    That is, the words, that when jointly removed, cause a substantial decline in the model's prediction confidence, exceeding a predefined threshold. 
    \item It assigns \textbf{importance scores} to each individual token, reflecting its impact on the final prediction. This score helps us understand how much each word contributes to the model's decision. 
    \item It provides \textbf{counterfactual explanations}, by showing the samples with minimal perturbation to the example, that lead to a different prediction. 
\end{enumerate}

\subsection{Setting and Notation}
\label{sec:setting-notation}
To summarize, our interest in this paper is directed toward explaining the prediction of a generic model $\model : \texts \to \mathcal{Y}$ (the \emph{black-box}) that takes textual documents as input. 
Note that in the case of a classification problem, $\model$ is a measurable function mapping textual inputs to confidence scores for $p$ different classes, \emph{i.e.}, $\mathcal{Y} = [0,1]^p$, and our goal then becomes to find the optimal subset of words that, if removed, significantly drops the confidence score with respect to the class of interest. 
Throughout this paper, we call $\doc$ a generic document, while $\xi$ denotes the specific document under consideration. 
We also define $\dictionary = \{\word_1,\ldots,\word_D\}$ as the \emph{global dictionary} containing~$D$ unique terms. 
Any document is a finite sequence of these dictionary elements. 
For a given example $\xi = (\xi_1,\ldots,\xi_b)$ composed of $b$ ordered words (not necessarily distinct), $\localdic_\xi = \{\word_1,\ldots,\word_d\} \in \dictionary$ captures the distinct words in $\xi$, with $d\leq b$. 
Additionally, $[k]$ represents the set of integers from~$1$ to~$k$. 

Finally, we define a \emph{candidate} explanation as any non-empty ordered sublist of $[b]$, corresponding to words of $\xi$. 
We call $\candidates$ the set of all candidates for $\xi$. 
We set $\size{\candidate}$ the length of the candidate, defined as the number of (not necessarily distinct) words that it contains. 

\subsection{Drop in prediction}
To assess the impact of removing specific words on model predictions, we introduce the key concept of \emph{drop in prediction}. 
For any sample $x$, we define this quantity as $\drop(x) \defeq \expec{\model(x)} - \model(x)$, where $\expec{\model(x)}$ is the expected prediction under the sampling distribution, and $x$ is a local perturbation of $\xi$ (detailed in Section~\ref{sec:sampling}). 
Subsequently, we characterize the drop of a candidate $\candidate$ as 
\begin{equation}
\label{eq:drop-candidate}
    \candrop_\candidate \defeq \expec{f(x)} - \expec{f(x) | \candidate \notin x} = 
    \expec{\drop(x) | \candidate \notin x} = \expecunder{\drop(x)}{c} \,,
\end{equation}
that is, the expected drop in prediction when a candidate is removed from the original document $\xi$. 
In essence, using Eq.~\eqref{eq:drop-candidate}, we attribute to any candidate the drop in prediction of samples where the candidate is perturbed. 
The optimal candidate, denoted as $\candidate^\star$, is determined by minimizing the size of the candidate subset while ensuring that it causes the average prediction $\expec{\model(x)}$ to drop by a significant amount, \emph{i.e.}, it is such that $\candrop_{\candidate^\star} \geq \epsilon \cdot \expec{\model(x)}$, as formulated by the optimization problem: 
\begin{align}
\label{eq:optimal-candiate}
\Minimize_{\substack{\candidate \in \candidates}} \size{\candidate} \,, \;  \text{subject to} \;\; \candrop_\candidate \geq \epsilon \cdot \expec{\model(x)}
\, .
\end{align}

\paragraph{Empirical drop in prediction.}
Calculating the prediction drop for each candidate in closed form, as formulated in Eq.~\eqref{eq:drop-candidate}, necessitates an exhaustive search and evaluation of $2^b$ candidates—an impractical endeavor for large documents. 
To circumvent this computational burden, we employ an empirical approach to estimate the prediction drop. 

For a given document $\xi$, we generate a set of $n$ samples through random perturbations of words in $\xi$ (detailed in Section~\ref{sec:sampling}). 
Consider any candidate~$\candidate$, and denote by $n_\candidate$ the quantity $\size{\{i \in [n] \mid \candidate \notin x_i\}}$, representing the number of samples where $\candidate$ is absent. 
In this context, we define the empirical drop of candidate $\candidate$ as 
\begin{equation}
\label{eq:empirical-drop}
    \Empdrop_\candidate \defeq \frac{1}{n} \sum_{i=1}^n \model(x_i) - \frac{1}{n_\candidate} \sum_{\candidate \notin x_i} \model(x_i) = \widehat{\model(x)}- \frac{1}{n_\candidate} \sum_{\candidate \notin x_i} \model(x_i)
    \, . 
\end{equation}
Note that the sampling scheme ensures that, \emph{with high probability}, for each candidate, there is at least one sample in which the candidate is not present. 

This definition guarantees that, for a large amount of samples, the empirical drop is a good estimate of Eq.~\eqref{eq:drop-candidate}, as expressed by the following: 

\begin{lemma}[Convergence of Empirical Drop $\Empdrop_\candidate$]
\label{lemma:empirical-drop}
For a candidate explanation $\candidate$, let $n_\candidate$ represent the count of instances in the dataset $x$ where $\candidate$ is not included in the sample. 
Then, as $n \to \infty$, the empirical drop in prediction $\Empdrop_\candidate$ associated to the candidate $\candidate$  converges in probability to 
\begin{align*}
    \expec{\model(x)} -\frac{\expec{\model(x) \indic{\candidate \notin x}}}{\proba{\candidate \notin x}} \,.
\end{align*}
\end{lemma}
This result motivates using Eq.~\eqref{eq:drop-candidate} instead of Eq.~\eqref{eq:empirical-drop} in subsequent analysis. 
Lemma~\ref{lemma:empirical-drop} is proved in Section~\ref{proof-lemma-empirical-drop} of the Appendix. 


\subsection{Sampling scheme}
\label{sec:sampling}
\begin{figure}[t]\resizebox{\textwidth}{!}{
    \centering
    \begin{minipage}{0.5\textwidth}
    \resizebox{\textwidth}{!}{
    \raggedleft 
    \setlength{\tabcolsep}{1.6pt}
    \begin{NiceTabular}{c | c c c c c c | c}
        & $\xi_1$ & $\xi_2$ & \Block[draw=red,rounded-corners]{9-1}{}\boldmath$\xi_3$ & $\xi_4$ & \Block[draw=red,rounded-corners]{9-1}{}\boldmath$\xi_5$ & $\xi_6$ & $\drop(x)$ \\ 
        \hline\hline
        $\xi$ & Poor & drinks & \textbf{decent} \color{black} & food & \textbf{great} \color{black} & service & $\drop(\xi)$ \\ 
        \hline
        $x_1$ & \color{red}{great} & drinks & decent & \color{red}{view} & \color{red}{slow} & service & $\drop(x_1)$ \\ 
        $x_2$ & Poor & \color{red}{seats} & \color{red}{bad} & \color{red}{boost} & great & \color{red}{house} & $\drop(x_2)$ \\ 
        \rowcolor{gray!15} $x_3$ & \color{red}{good} & \color{red}{table} & \color{red}{poor} & food & \color{red}{awful} & service & $\drop(x_3)$ \\ 
        $x_4$ & \color{red}{amazing} & \color{red}{spot} & decent & food & \color{red}{bad} & \color{red}{tips} & $\drop(x_4)$ \\ 
        \rowcolor{gray!15} $x_5$ & Poor & drinks & \color{red}{boring} & \color{red}{walk} & \color{red}{inept}  & service & $\drop(x_5)$ \\ 
        $\vdots$ & $\vdots$ & $\vdots$ & $\vdots$ & $\vdots$ & $\vdots$ & $\vdots$ & $\vdots$ \\ 
        \rowcolor{gray!15}  $x_n$ & Poor & \color{red}{space} & \color{red}{average} & food & \color{red}{lousy} & service & $\drop(x_n)$ \\ 
        \end{NiceTabular}} 
    \end{minipage}
    \hfill
    \begin{minipage}{0.5\textwidth}    
    \resizebox{\textwidth}{!}{
    \raggedright 
    \setlength{\tabcolsep}{2.8pt}
    \begin{NiceTabular}{c | c c c c c c | c}
        & $\xi_1$ & $\xi_2$ &  \Block[draw=red,rounded-corners]{9-1}{}\boldmath$\xi_3$ & $\xi_4$ & \Block[draw=red,rounded-corners]{9-1}{}\boldmath$\xi_5$ & $\xi_6$ & $\drop(x)$ \\ 
        \hline\hline
        $\xi$ & Poor & drinks & \textbf{decent} \color{black} & food & \textbf{great} \color{black} & service & $\drop(\xi)$ \\ 
        \hline
        $x_1$ & \color{red}\texttt{UNK} & drinks & decent & \color{red}\texttt{UNK} & \color{red}\texttt{UNK} & service & $\drop(x_1)$ \\ 
        $x_2$ & Poor & \color{red}\texttt{UNK} & \color{red}\texttt{UNK} & \color{red}\texttt{UNK} & great & \color{red}\texttt{UNK} & $\drop(x_2)$ \\ 
        \rowcolor{gray!15} $x_3$ & \color{red}\texttt{UNK} & \color{red}\texttt{UNK} & \color{red}\texttt{UNK} & food & \color{red}\texttt{UNK} & service & $\drop(x_3)$ \\ 
        $x_4$ & \color{red}\texttt{UNK} & \color{red}\texttt{UNK} & decent & food & \color{red}\texttt{UNK} & \color{red}\texttt{UNK} & $\drop(x_4)$ \\ 
        \rowcolor{gray!15} $x_5$ & Poor & drinks & \color{red}\texttt{UNK} & \color{red}\texttt{UNK} & \color{red}\texttt{UNK}  & service & $\drop(x_5)$ \\ 
        $\vdots$ & $\vdots$ & $\vdots$ & $\vdots$ & $\vdots$ & $\vdots$ & $\vdots$ & $\vdots$ \\ 
        \rowcolor{gray!15} $x_n$ & Poor & \color{red}\texttt{UNK} & \color{red}\texttt{UNK} & food & \color{red}\texttt{UNK} & service & $\drop(x_n)$ \\ 
        \end{NiceTabular}
    }
    \end{minipage}
}
\caption{Illustration of FRED's \texttt{pos-sampling} scheme (left panel) and \texttt{mask-sampling} scheme (right panel) for computing the drop of a candidate. 
For a given example $\xi$, FRED generates $n$ perturbed samples $x_1, \ldots, x_n$ by independently perturbing tokens with probability $p(=0.5)$. Each sample is associated with the model's drop in prediction $\drop(x_j)$. 
Finally, the empirical drop $\Empdrop_{\candidate}$ of a candidate is computed by averaging the drops over the samples that do not contain $\candidate$. 
In the example, the candidate consists of the words \emph{decent} and \emph{great}. 
The samples where both tokens are perturbed are highlighted in gray. 
The empirical drop associated to \{\emph{decent}, \emph{great}\} is therefore computed by averaging $\drop(x_3)$, $\drop(x_5)$, $\ldots$ $\drop(x_n)$.}
\label{fig:sampling-scheme}
\end{figure}

We now detail the sampling scheme used to estimate the drop associated to a candidate (see Eq.~\eqref{eq:drop-candidate}). 
Again, the goal is to look at the behavior of the model~$\model$ in a local neighborhood of $\xi$, \emph{i.e.}, when some words in $\xi$ are absent. 
To avoid out-of-distribution samples, we replace absent tokens by random
words with the same Part-of-Speech (POS) tag \citep{ribeiro2018anchors}. 
This means that, for instance, a verb will be replaced by another verb, and an adjective with another adjective. 
Additionally, to highlight the difference in prediction, if the absent token has a \emph{sentiment} (positive or negative), FRED replaces it with a word with same POS tag and opposite sentiment. 
Indeed, in the example of Figure~\ref{fig:sampling-scheme}, replacing the word \say{great} with the word \say{amazing,} will probably not result in any significant change in prediction. 
Contrarily, replacing it with \say{bad} or \say{awful} will highlight its impact. 
We refer to this approach as \textbf{pos-sampling}. 
Specifically, given a corpus $\texts$, FRED under the \texttt{pos-sampling} scheme creates two dictionaries $\dictionary_{pos}^+$ and $\dictionary_{pos}^-$ corresponding to the set of tokens having the same POS tag with \emph{positive} and \emph{negative} sentiment (\emph{neutral} words are in both sets). 
Then, for a given example $\xi$, FRED generates perturbed samples $x_1, \ldots, x_n$ as follows: 
\begin{enumerate}
    \item $n$ copies of $\xi$ are created; 
    \item for each sample, each token is independently selected with probability $p$;
    \item selected tokens are associated with a pair (\emph{pos}, \emph{sentiment}); 
    \item selected tokens are replaced with a random (uniformly chosen) word with same \emph{pos} and opposite \emph{sentiment} from $\dictionary_{pos}^+$ and $\dictionary_{pos}^-$, accordingly; 
    \item each sample $x_j$ is associated with its drop in prediction $\drop(x_j)$. 
\end{enumerate}
See Figure~\ref{fig:sampling-scheme} for an illustration. 

The sample size $n$ is chosen such that, for each candidate, there exists with high probability at least one sample not containing it, \emph{i.e.}, such that for each $\candidate \in \candidates$, $n_\candidate \geq 1$ with probability higher than $\alpha$ (see Lemma~\ref{lemma:n-sample}). 

\begin{lemma}[Choosing $n$]
\label{lemma:n-sample}
If the sample size $n$ is higher or equal than the quantity $\frac{\logs{1 - \alpha}}{\logs{1 - 1/2^{\ell_{max}}}}$, then, for any candidate $\candidate$ with size smaller or equal than $\length_{max}$, there exists at least one sample not containing it, with probability higher or equal than $\alpha$, \emph{i.e.}, for any candidate~$\candidate$
\[
n = \ceil{\frac{\logs{1 - \alpha}}{\logs{1 - 1/2^{\ell_{max}}}}} \Longrightarrow \proba{\exists i \in [n] : \candidate \notin x_i} \geq \alpha 
\,.
\]
\end{lemma}
As \say{high probability,} we set by default $\alpha = 0.95$ and the token perturbation probability $p=0.5$. 
As a maximum number of words to be used as an explanation, we set $\ell_{max} = 10$: we realistically believe it is not helpful to use a high proportion of a text. 
According to Lemma~\ref{lemma:n-sample} (proven in Appendix~\ref{proof-lemma-n-sample}), these choices imply a sample size of $n \approx 3000$. 

\paragraph{Remark.} 
Note that this sampling scheme requires the availability of a corpus $\texts$, and can be computationally costly, as it requires the computation of the POS-tag and the sentiment for any token. 
Additionally, its effectiveness can depend on the underlying model. 
Alternatively, we follow the default implementation of the official repositories from Anchor, LIME, and SHAP, and propose the \textbf{mask-sampling}: simply mask the tokens to be removed with a predefined token. 

The \texttt{mask-sampling} scheme is often criticized for creating out-of-distribution samples \citep{hase2021out} by generating meaningless sentences. 
To address this issue, Anchors' official implementation has the option of sampling by making BERT predict the missing words. 
Such a solution is potentially interesting, however, we believe that in practice it does not meet expectations. 
First, replacing words with large language models is extremely resource intensive: \texttt{BERT}-Anchor is $10$ times slower than \texttt{mask}-Anchor, even on small documents \citep{lopardo2022sea}, with no significant advantage in explanations. 
Second, by examining the generated samples, we observe that the sentences are often still meaningless. 
Note that \citet{ribeiro2018anchors} proposes replacing the tokens with random words with the same part-of-speech tag with probability proportional to their similarity in the embedding space. 
As already stated, we find the opposite-sentiment approach to be most suitable for our purposes: it better captures the impact of the original token. 

\subsection{Explanations} 
\label{sec:fred-exp} 
Once the sample $x$ is created, the empirical drop associated to a candidate $\Empdrop_{\candidate}$ is computed by averaging the drops over the samples that do not contain $\candidate$. 
See Figure~\ref{fig:sampling-scheme} for an illustration. 
At this point, the \textbf{minimal influential subset} of tokens is identified by the empirical version of Eq.~\eqref{eq:optimal-candiate}:
\begin{align}
\label{eq:optimal-emp-candiate}
\Minimize_{\substack{\candidate \in \candidates}} \size{\candidate} \,, \;  \text{subject to} \;\; \Empdrop_\candidate \geq \epsilon \cdot \widehat{\model(x)}
\, .
\end{align}
Algorithm~\ref{algo:fred} illustrates FRED's mechanism, and its output is shown in Figure~\ref{fig:fred-exp}. 
\alglanguage{pseudocode}
\begin{algorithm}[t]
\caption{\label{algo:fred}An overview of FRED's algorithm. By default, we set $\epsilon=0.15$, $\ell_{max} = 10$, $p=0.5$, and compute the sample size $n$ according to Lemma \ref{lemma:n-sample}.}
\begin{algorithmic}
\State \textbf{input} model $\model$, example $\xi$, threshold $\epsilon$, max\_len $\ell_{max}$, p\_perturb $p$, n\_sample $n$
\State \textbf{initialize} $\Empdrop_{best} = 0$
\State $x$ = \texttt{generate\_sample}($\xi$, $p$, $n$) \Comment{according to \texttt{pos-sampling} or \texttt{mask-sampling}}      
\State $\widehat{\model(x)} = \texttt{average}(\model(x))$
\State $\drop(x) = \widehat{\model(x)} - \model(x)$  \Comment{difference between average and each sample prediction}
\For{\texttt{size} = 1: $\ell_{max}$}: 
    \State \texttt{candidates} = \texttt{get\_candidates}($\xi$, \texttt{size})   \Comment{return subsets with cardinality \texttt{size}}
    \For{$\candidate$ in \texttt{candidates}}:
        \State $\Empdrop_\candidate$ = \texttt{compute\_drop}($\model$, $\candidate$, $x$, $\drop(x)$) \Comment{average $\drop(x)$ where $c\notin x$} 
        \If{size = $1$}:
            \State $s_c = \Empdrop_\candidate$ \Comment{per-token importance score}      
        \EndIf      
        \If{$\Empdrop_\candidate \geq \Empdrop_{best}$}: 
            \State $\Empdrop_{best} = \Empdrop_{\candidate}$, \texttt{best} = $\candidate$
        \EndIf
    \EndFor
    \If{$\Empdrop_{best} \geq \epsilon \widehat{\model(x)}$}:
        \State \textbf{return} \texttt{best} 
    \EndIf
\EndFor
\State \textbf{return} \texttt{best} 
\end{algorithmic}
\end{algorithm} 
Note that the candidates of size $1$ correspond to the tokens in the example. 
Therefore, FRED assigns \textbf{importance score} to each token in the example $\xi$ as 
\begin{align} 
\label{eq:token-score} 
\forall i \in [b]\,, \quad s_i = \Empdrop_{\{i\}}
\,. 
\end{align}
In practice, the importance score $s_i$ is the average drop of samples where the $i$-th token is perturbed. 
Figure~\ref{fig:fred-exp} (b) shows the saliency maps for such scores. 

Additionally, for classification tasks, FRED provides \textbf{counterfactual explanations} by identifying the samples in $x$ with minimal perturbation (\emph{i.e.}, with minimal amount of perturbed tokens), leading to a different classification. 
See Figure~\ref{fig:fred-exp} (c) for an illustration. 

\section{Analysis on Explainable Classifiers}
\label{sec:analysis}
To rigorously assess FRED's effectiveness, we perform a comprehensive theoretical analysis. 
We leverage the framework established by \citet{garreau2020explaining} for LIME, extended for text data by \citet{mardaoui_garreau_2021,lopardo2022sea}. 
Since there is no perfect way to verify explanation correctness, we ensure FRED behaves as expected on simple, well-understood models.
Our investigation focuses on FRED's performance with inherently interpretable classifiers. 
Firstly, we examine its ability to identify the most critical token subset when applied to linear models.
Secondly, we evaluate its capability in detecting shortcuts \citep{bastings2021will}. 

For our theoretical analysis, we make two key assumptions about the models we examine. 
First, we focus on binary sentiment analysis tasks in this section. 
We assume, without loss of generality, that the example $\xi$ is classified as positive. 
In other words, the model $\model$ outputs a confidence score for the positive class. 
Second, to simplify the analysis and obtain closed-form solutions, we consider classifiers that operate on the well-established TF-IDF (Term Frequency-Inverse Document Frequency) \citep{luhn1957statistical} vectorization $\Tfidf$ of documents. 
While not the most recent technique, TF-IDF's simplicity allows us to derive closed-form solutions for our analysis. 
We denote by $\mult_j(\doc)$ the multiplicity of word $\word_j$ in $\doc$, \emph{i.e.},
\[
\mult_j(\doc) \defeq \card{\{k \in [b], \, \doc_k = \word_j\}}
\, .
\]
When the context is clear, we write $\mult_j$ short for $\mult_j(\xi)$. 
\begin{definition}[TF-IDF]
\label{def:tf-idf}
Let $N$ be the size of the initial corpus $\texts$, \emph{i.e.}, the number of documents in the dataset. 
Let $N_j$ be the number of documents containing the word $\word_j \in \dictionary$. 
The TF-IDF of $\doc$ is the vector $\tfidf{\doc} \in \Reals^D$ such that
\[
\forall j \in [D], \qquad\tfidf{\doc}_j \defeq \mult_j(\doc) \idf_j 
\,,
\]
where $\idf_j \defeq \log{\frac{N + 1}{N_j + 1}} + 1$ is the \emph{inverse document frequency} (IDF) of $\word_j$ in  $\texts$.
\end{definition}

We call a \emph{candidate} explanation $\candidate$ any ordered sublist of words in $\xi$, formally defined as $\candidate = (\candidate_1, \candidate_2, \ldots, \candidate_d)$, where $0 \leq \candidate_j \leq \mult_j$ for all $ j \in [d]$. 
We define the length $\length$ of a candidate $\length = \size{\candidate} = \sum_{j=1}^d \candidate_j$. 

Note that once the TF-IDF vectorizer is fitted on a corpus, the vocabulary is fixed: if a word is not part of the initial corpus, its IDF term is zero. 
We make the (realistic) assumption that the mask token is not in the fitted corpus and consider the \texttt{mask}-sampling seen in Section~\ref{sec:sampling}. 
Thus, replacing any word with this token is equivalent to simply removing it, from the point of view of TF-IDF. 

\subsection{Linear classifiers}
\label{sec:linear-classifiers}
Let us now direct our attention towards the exploration of \emph{linear classifiers}, that is, classifiers whose underlying models are defined as
\begin{equation}
\label{eq:def-linear-classifier}
\model(\doc) = \lambda^\top \tfidf{\doc} + \lambda_0 \,,
\end{equation}
where $\lambda \in \Reals^D$ represents a learned vector of coefficients, and $\lambda_0 \in \Reals$ serves as an intercept term. 
This linear classifier setup encapsulates a broad range of models, including logistic models and perceptron, making it an insightful starting point for our analysis. 
The decision boundary of the classifier is determined by the hyperplane where the linear combination of the TF-IDF vector $\tfidf{\doc}$, as weighted by $\lambda$, along with the intercept $\lambda_0$, is greater than zero. 
\begin{proposition}[Linear models]
\label{prop:linear-models}
Let $\lambda,\lambda_0$ be the coefficients associated to the linear classifier defined by Eq.~\eqref{eq:def-linear-classifier}. 
Assume $\lambda_1 \idf_1 > \lambda_2 \idf_2 > \cdots > \lambda_d \idf_d$. 
Then, the solution to Eq.~\eqref{eq:optimal-candiate} is such that (i) words associated with a negative coefficient do not appear in the optimal candidate, and (ii) the algorithm starts including the words with higher $\lambda_j\idf_j$ until a threshold is met. 
\end{proposition}

In simpler terms, Proposition~\ref{prop:linear-models} reveals that {in the case of a linear classifier, FRED retains only words that positively impact the prediction}. 
Furthermore, it {prioritizes words with the highest $\lambda_j\idf_j$ values}, gradually including them in the explanation until the desired confidence level is achieved.
An illustration is shown in Figure~\ref{fig:linear-illustation}. 
We provide a proof of Proposition~\ref{prop:linear-models} in Appendix~\ref{proof-prop-linear-models}. 

\begin{figure}[t]
    \centering
    \begin{tikzpicture}
    \draw[<-] (0,-0.5) -- (0,6) node[left, midway, rotate=90, yshift=1.0cm, xshift=3.0cm] {$\Empdrop_\candidate \approx \expec{\model(x)} - \expecunder{\model(x)}{c}$};

    \draw (-0.1,5.7) -- (0.1,5.7) node [right] {$(1,0,0,0,0,\ldots,0)$};
    \draw (-0.1,5.0) -- (0.1,5.0) node [right] {$(2,0,0,0,0,\ldots,0)$};
    \draw (0.0, 4.35) node [right, xshift=0.4cm] {$\vdots$};
    \draw (-0.1,3.7) -- (0.1,3.7) node [right] {$(\mult_1,\mult_2,0,0,0,\ldots,0)$};
    \draw (-0.1,3.0) -- (0.1,3.0) node [right] {$(\mult_1,\mult_2,1,0,0,\ldots,0)$};
    \draw (0.0, 2.35) node [right, xshift=0.4cm] {$\vdots$};
    \draw (-0.1,1.5) -- (0.1,1.5) node [right] {$(\mult_1,\mult_2,\mult_3,1,0,\ldots,0)$};
    \draw[blue] (-0.1,0.5) -- (0.1,0.5) node [right] {$(\mult_1,\mult_2,\mult_3,2,0,\ldots,0)$};
    \draw (-0.1,0.0) -- (0.1,0.0) node [right] {$(\mult_1,\mult_2,\mult_3,3,0,\ldots,0)$};

    \draw[red, line width=1.5pt] (-0.2,0.9) node [left] {$\varepsilon \cdot \expec{\model(x)}$} -- (4.0,0.9);

\end{tikzpicture}
    \caption{\label{fig:linear-illustation}Illustration of Proposition~\ref{prop:linear-models}. On linear models, Algorithm \ref{algo:fred} includes words having the highest $\lambda_j\idf_j$s first. Finally, the minimal candidate satisfying the threshold condition is selected, which is $\candidate=(\mult_1,\mult_2,\mult_3,2,0,\ldots,0,0)$ in the example.}
\end{figure}
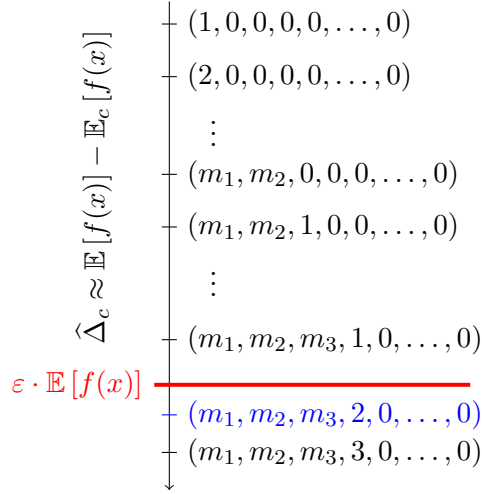

\subsection{Shortcuts detection}
\label{sec:rules}
Moving forward, our attention shifts to classifiers that are grounded in the existence or nonexistence of specific tokens (shortcuts) in a document. 
This consideration stems from our utilization of the TF-IDF vectorizer, where the presence or absence of a word $\word_i$ in a document $\doc$ is straightforwardly captured by the condition $\tfidf{\doc}_i > 0$ (or $\tfidf{\doc}_i = 0$). 
This characterization allows us to explore classifiers whose predictions hinge upon the occurrence or non-occurrence of particular terms in the document. 
\begin{proposition}[Presence of shortcuts]
\label{prop:presence-words}
Let us assume that $\mult_1 < \mult_2 < \ldots < \mult_k$, where $k$ is the maximum number of unique words in the document.
Consider the set $J = \{1, 2, \ldots, k\}$ as a subset of $[d]$, and suppose that the model is defined as
\[
\model(\doc) = \indic{\word_j \in \doc,\; \forall j \in J} = \prod_{j \in J} \indic{\word_j \in \doc} = \prod_{j \in J} \indic{\tfidf{\doc}_j > 0}.
\]
Then, the resulting optimal candidate $\candidate^\star$ is characterized by $\candidate^\star_1 = \min\{\mult_1, \length_{max}\}$ and $\candidate^\star_j = 0$ for $j \geq 2$. 
\end{proposition}

In essence, if a model classifies a document based on the presence of specific words, removing all occurrences of just one of those words is sufficient to change the classification. 
FRED captures this information by identifying the minimal set of words whose removal would trigger this change. 
Proposition~\ref{prop:presence-words} pinpoints that the smallest set of words necessary to trigger this shift in prediction is composed of the word with the least number of occurrences in the document. 
This is a reassuring property of FRED that aligns well with our definition of explanations. 
Proposition~\ref{prop:presence-words} is proven in Appendix~\ref{proof-prop-presence-words}. 

\section{Experiments}
\label{sec:experiments}
We assess FRED's explanation quality against popular methods (LIME \citep{ribeiro2016should}, SHAP \citep{lundberg2017unified}, Anchors \citep{ribeiro2018anchors}) focusing on \textbf{faithfulness}, \emph{i.e.}, the adherence of the explanation to the model's behavior, and its \textbf{robustness}. 
We evaluate on three sentiment analysis datasets (Restaurants, Yelp reviews, IMDb) and a hate speech detection dataset (Tweets) with varying document lengths (details in Appendix~\ref{appendix-sec:experiments}). 
We trained logistic regression, decision trees, and random forests on each dataset. 
Additionally, we applied pre-trained RoBERTa \citep{liu2019roberta} and DistilBERT \citep{sanh2019distilbert} models. 

\paragraph{Faithfulness.}
Faithfulness is evaluated using \textbf{Comprehensiveness} and \textbf{Sufficiency} (as in \citet{deyoung2020eraser}) for evaluating explainers' ability to identify the crucial token subset, and \textbf{AUC-MoRF} (Area Under the Most Relevant First Perturbation Curve) \citep{kakogeorgiou2021evaluating} for the importance-based explanations.

Given an explanation $\explanation$ defined as a subset of words in the example $\xi$ to explain, the \textbf{Comprehensiveness} is computed as the difference between the prediction confidence of the entire document $\xi$ and the prediction confidence when the explanation $\explanation$ is removed. 
The \textbf{Sufficiency} is computed as the difference between the prediction confidence of the entire document $\xi$ and the prediction confidence based only on the explanation $\explanation$. 
Thus, 
\[ \text{Comprehensiveness} = \model(\xi) - \model(\xi \setminus \explanation) \qquad \text{and} \qquad \text{Sufficiency} = \model(\xi) - \model(\explanation) \,. \]
A high Comprehensiveness score implies that the subset $\explanation$ was indeed influential in the prediction, while a low score suggests that it had little impact on the prediction outcome. 
A low Sufficiency score implies that the subset $\explanation$ does not sufficiently summarize the document, and more information is needed to make an accurate prediction. 

When applying feature-importance based explainers, we can rank the tokens in $\xi$ according to their score. 
The AUC-MoRF can be defined as 
\[ \text{AUCMoRF} = \frac{1}{D} \sum_{k=2}^{D} \frac{f(y^{(k-1)}) + f(y^{(k)})}{2} \,, \] 
where $D$ is the maximum number of perturbation, and $\mathbf{y}^{(k)}$ is the example $\xi$ after the $k^{\text{th}}$ MoRF perturbation, \emph{i.e.}, after removing the $k$ most important (positive) tokens according to the explanation. 
A smaller value of AUC-MoRF means a more
faithful explanation. 
We stress out that only the tokens with positive score are perturbed for this computation. 
Let us call $\explanation^+$ the list of tokens with positive influence according to the explainer in exam. 
In our experiments, we set $D = \min{(20, \size{\explanation^+})}$. 

For the faithfulness metrics, token removals are simulated by replacing the missing token with the mask \texttt{UNK}.

\paragraph{Robustness.} 
We compute the \textbf{Robustness} of an explanation $\explanation$ as follows. 
First, we apply the explainer, retaining its explanation $\explanation$ as the ground truth for the example. 
Then, we conduct $k$ additional iterations of the explainer on the same document, yielding $k$ new explanations, denoted as $e_1, e_2, \ldots, e_k$ for that instance. 
We therefore compute Jaccard Similarity between the original explanation $\explanation$ and each $e_i$, $i \in [k]$. 
Finally, we compute the Robustness as the average Jaccard Similarity scores across the $k$ new explanations and the original explanation. 
In practice, 
\[ 
\forall i \in [k] \,,\: \text{J}(e, e_i) = \frac{{|e \cap e_i|}}{{|e \cup e_i|}} \quad \text{and} \quad \text{Robustness} = \frac{{\sum_{{i=1}}^{k} \text{J}(e, e_i)}}{{k}} \,. \] 

\begin{table}[t]
\caption{\label{exp-tab:restaurants_roberta_5}
Comparison on Roberta for Restaurant reviews ($p=0.5$, $\epsilon=0.15$).}
\centering
\setlength{\tabcolsep}{2pt}
\begin{tabular}{@{}l|rrrrrrr@{}}
\textbf{} & \texttt{suffic.} $\downarrow$ & \texttt{compreh.} $\uparrow$ & \texttt{robust}. $\uparrow$ & \texttt{aucmorf} $\downarrow$ & \texttt{time (s)} $\downarrow$ & \texttt{proport.} $\downarrow$ & \\
\hline
fred & $0.517 (0.49)$ & \underline{$0.548 (0.49)$} & $0.925 (0.22)$ & \underline{$0.146 (0.11)$} & $35.814 (1.69)$ & $0.127 (0.06)$ & \\
fredpos & $0.517 (0.49)$ & $0.542 (0.49)$ & $0.920 (0.23)$ & $0.198 (0.16)$ & $35.249 (1.57)$ & $0.127 (0.06)$ & \\
lime & $0.527 (0.49)$ & $0.528 (0.50)$ & $0.925 (0.22)$ & $0.162 (0.13)$ & $38.834 (2.72)$ & $0.127 (0.06)$ & \\
shap & $0.766 (0.42)$ & $0.323 (0.46)$ & \underline{$1.000 (0.00)$} & $0.306 (0.21)$ & \underline{$1.899 (1.74)$} & $0.127 (0.06)$ & \\
anchor & \underline{$0.509 (0.49)$} & $0.538 (0.50)$ & $0.850 (0.32)$ & $0.454 (0.45)$ & $15.491 (20.40)$ & \underline{$0.126 (0.06)$} & \\
\end{tabular}
\end{table}

\begin{table}[t]
\caption{\label{exp-tab:yelp_forest_5}
Comparison on Random forest classifier for Yelp reviews ($p=0.5$, $\epsilon=0.15$).}
\centering
\setlength{\tabcolsep}{2pt}
\begin{tabular}{@{}l|rrrrrrr@{}}
\textbf{} & \texttt{suffic.} $\downarrow$ & \texttt{compreh.} $\uparrow$ & \texttt{robust}. $\uparrow$ & \texttt{aucmorf} $\downarrow$ & \texttt{time (s)} $\downarrow$ & \texttt{proport.} $\downarrow$ & \\
\hline
fred & \underline{$-0.142 (0.11)$} & \underline{$0.114 (0.04)$} & $0.804 (0.21)$ & $0.759 (0.13)$ & $0.418 (0.18)$ & $0.117 (0.14)$ & \\
fredpos & $-0.123 (0.12)$ & $0.080 (0.05)$ & $0.833 (0.23)$ & \underline{$0.740 (0.09)$} & $0.452 (0.21)$ & $0.071 (0.10)$ & \\
lime & $-0.126 (0.11)$ & $0.083 (0.04)$ & $0.933 (0.16)$ & $0.782 (0.08)$ & \underline{$0.332 (0.10)$} & $0.061 (0.08)$ & \\
shap & $-0.132 (0.11)$ & $0.073 (0.05)$ & \underline{$0.972 (0.09)$} & $0.776 (0.08)$ & $0.638 (0.23)$ & $0.071 (0.10)$ & \\
anchor & $-0.049 (0.16)$ & $0.032 (0.05)$ & $0.754 (0.33)$ & $0.942 (0.09)$ & $2.530 (6.26)$ & \underline{$0.038 (0.04)$} & \\
\end{tabular}
\end{table}

\begin{table}[t]
\caption{\label{exp-tab:imdb_distilbert_5}
Comparison on DistilBERT for IMDb ($p=0.5$, $\epsilon=0.15$).}
\centering
\setlength{\tabcolsep}{2pt}
\begin{tabular}{@{}l|rrrrrrr@{}}
\textbf{} & \texttt{suffic.} $\downarrow$ & \texttt{compreh.} $\uparrow$ & \texttt{robust}. $\uparrow$ & \texttt{aucmorf} $\downarrow$ & \texttt{time (s)} $\downarrow$ & \texttt{proport.} $\downarrow$ & \\
\hline
fred & \underline{$-0.002 (0.01)$} & \underline{$0.020 (0.01)$} & $0.602 (0.15)$ & \underline{$0.970 (0.01)$} & $9.406 (0.87)$ & $0.387 (0.13)$ & \\
fredpos & $0.002 (0.01)$ & $0.017 (0.01)$ & $0.499 (0.14)$ & $0.975 (0.01)$ & $9.754 (0.65)$ & $0.381 (0.13)$ & \\
lime & $0.001 (0.01)$ & $0.019 (0.01)$ & $0.897 (0.10)$ & $0.972 (0.01)$ & $11.052 (1.31)$ & $0.292 (0.15)$ & \\
shap & $0.000 (0.01)$ & $0.016 (0.01)$ & \underline{$1.000 (0.00)$} & $0.975 (0.01)$ & $1.934 (0.86)$ & $0.381 (0.13)$ & \\
anchor & $0.020 (0.01)$ & $0.003 (0.00)$ & \underline{$1.000 (0.00)$} & $0.995 (0.01)$ & \underline{$0.633 (0.06)$} & \underline{$0.039 (0.02)$} & \\
\end{tabular}
\end{table}

\begin{table}
\caption{\label{exp-tab:imdb_roberta_5}
Comparison on Roberta for IMDb ($p=0.5$, $\epsilon=0.15$).}
\centering
\setlength{\tabcolsep}{2pt}
\begin{tabular}{@{}l|rrrrrrr@{}}
\textbf{} & \texttt{suffic.} $\downarrow$ & \texttt{compreh.} $\uparrow$ & \texttt{robust}. $\uparrow$ & \texttt{aucmorf} $\downarrow$ & \texttt{time (s)} $\downarrow$ & \texttt{proport.} $\downarrow$ & \\
\hline
fred & \underline{$0.201 (0.39)$} & \underline{$0.249 (0.43)$} & $0.864 (0.23)$ & \underline{$0.204 (0.17)$} & $47.849 (4.14)$ & $0.069 (0.05)$ & \\
fredpos & $0.454 (0.49)$ & $0.209 (0.40)$ & $0.910 (0.27)$ & $0.328 (0.26)$ & $48.098 (4.51)$ & $0.038 (0.02)$ & \\
lime & $0.348 (0.46)$ & $0.219 (0.41)$ & $0.810 (0.37)$ & $0.295 (0.33)$ & $66.732 (8.62)$ & $0.038 (0.02)$ & \\
shap & $0.621 (0.47)$ & $0.149 (0.35)$ & \underline{$1.000 (0.00)$} & $0.475 (0.32)$ & \underline{$15.147 (4.37)$} & $0.038 (0.02)$ & \\
anchor & $0.463 (0.48)$ & $0.228 (0.42)$ & $0.640 (0.41)$ & $0.777 (0.40)$ & $55.617 (115.33)$ & \underline{$0.037 (0.02)$} & \\
\end{tabular}
\end{table}

\begin{table}
\caption{\label{exp-tab:tweets_tree_5}
Comparison on a decision tree for Tweets ($p=0.5$, $\epsilon=0.15$).}
\centering
\setlength{\tabcolsep}{2pt}
\begin{tabular}{@{}l|rrrrrrr@{}}
\textbf{} & \texttt{suffic.} $\downarrow$ & \texttt{compreh.} $\uparrow$ & \texttt{robust}. $\uparrow$ & \texttt{aucmorf} $\downarrow$ & \texttt{time (s)} $\downarrow$ & \texttt{proport.} $\downarrow$ & \\
\hline
fred & \underline{$0.870 (0.34)$} & \underline{$0.930 (0.26)$} & $0.975 (0.12)$ & $0.079 (0.04)$ & $0.054 (0.01)$ & $0.140 (0.05)$ & \\
fredpos & $0.880 (0.32)$ & \underline{$0.930 (0.26)$} & $0.954 (0.14)$ & \underline{$0.078 (0.04)$} & $0.065 (0.01)$ & $0.140 (0.05)$ & \\
lime & \underline{$0.870 (0.34)$} & $0.920 (0.27)$ & $0.881 (0.22)$ & $0.082 (0.06)$ & $0.129 (0.01)$ & $0.140 (0.05)$ & \\
shap & $0.880 (0.32)$ & $0.910 (0.29)$ & \underline{$1.000 (0.00)$} & $0.088 (0.08)$ & \underline{$0.034 (0.16)$} & $0.140 (0.05)$ & \\
anchor & $0.890 (0.31)$ & $0.800 (0.40)$ & $0.650 (0.39)$ & $0.109 (0.14)$ & $0.582 (0.48)$ & \underline{$0.138 (0.05)$} & \\
\end{tabular}
\end{table}

\begin{table}
\caption{\label{exp-tab:tweets_forest_5}
Comparison on random forest classifier for Tweets ($p=0.5$, $\epsilon=0.15$).}
\centering
\setlength{\tabcolsep}{2pt}
\begin{tabular}{@{}l|rrrrrrr@{}}
\textbf{} & \texttt{suffic.} $\downarrow$ & \texttt{compreh.} $\uparrow$ & \texttt{robust}. $\uparrow$ & \texttt{aucmorf} $\downarrow$ & \texttt{time (s)} $\downarrow$ & \texttt{proport.} $\downarrow$ & \\
\hline
fred & \underline{$0.781 (0.21)$} & $0.295 (0.19)$ & $0.892 (0.21)$ & $0.156 (0.05)$ & $0.512 (0.04)$ & $0.108 (0.03)$ & \\
fredpos & $0.784 (0.20)$ & \underline{$0.386 (0.15)$} & $0.958 (0.15)$ & \underline{$0.149 (0.05)$} & $0.379 (0.05)$ & $0.108 (0.03)$ & \\
lime & $0.784 (0.20)$ & $0.384 (0.15)$ & $0.974 (0.14)$ & $0.159 (0.05)$ & $0.452 (0.05)$ & $0.108 (0.03)$ & \\
shap & $0.784 (0.20)$ & $0.383 (0.15)$ & $1.000 (0.00)$ & $0.156 (0.05)$ & \underline{$0.155 (0.15)$} & $0.108 (0.03)$ & \\
anchor & $0.788 (0.20)$ & $0.281 (0.19)$ & $0.493 (0.40)$ & $0.221 (0.10)$ & $8.019 (4.09)$ & $0.108 (0.03)$ & \\
\end{tabular}
\end{table}

\paragraph{Results.}
Along with the metrics above, we report here the average computing time, and the average proportion of document (\emph{i.e.}, $\size{\explanation} / \size{\xi}$) used for explanation (note that AUC-MoRF is independent of this). 
We refer below FRED under the \texttt{pos-sampling} scheme as FRED-pos (\texttt{fredpos} in the tables), and to FRED under the \texttt{mask-sampling} as FRED-mask (\texttt{fred} in the tables). 
We report here the comparison on 
Roberta for Restaurants reviews (Table~\ref{exp-tab:restaurants_roberta_5}, $k=2$ for Robustness), 
random forest classifier for Yelp reviews (Table~\ref{exp-tab:yelp_forest_5}, $k=10$ for Robustness), 
DistilBERT and Roberta on IMDb (Tables~\ref{exp-tab:imdb_distilbert_5} and \ref{exp-tab:imdb_roberta_5}, $k=2$ for Robustness),
and
decision tree and random forest classifier for Tweets hate speech detection (Tables~\ref{exp-tab:tweets_tree_5} and \ref{exp-tab:tweets_forest_5}, $k=10$ for Robustness). 

Both FRED-mask and FRED-pos produce significantly more faithful explanations than the others. 
FRED-mask is even slightly better than FRED-pos in terms of sufficiency and comprehensiveness, but note that it requires a larger number of tokens (proportion is higher). 
In general, Anchor performs well on small documents (Restaurant dataset), while its behavior on larger documents is highly nonlinear, as it tends to be conservative on the size of the anchor. 
We also note that it behaves poorly only documents classified with a high confidence (as proved \citep{lopardo2022sea}). 
LIME and SHAP have a close behavior, as somehow expected, but the latter is significantly more efficient. 
On small documents, SHAP performs an exhaustive search over all possible token subsets, motivating its high robustness. 
Due to space constraint, we refer to Appendix~\ref{appendix-sec:experiments} for additional details on the experimental setting and results. 
%


\section{Conclusion}
\label{sec:conclusion}
Building trustworthy AI necessitates interpretable machine learning, especially in critical domains. 
Existing explainers for text models, however, often grapple with complexity, lack formal grounding, and unreliable performance. 
Our proposed method, FRED, tackles these limitations by providing three key explanatory insights: 1) identifying the minimal set of crucial words, 2) assigning importance scores to each token, and 3) generating counterfactual explanations. 
We formally established FRED's reliability through theoretical analysis on interpretable models, while our empirical evaluation assess its effectiveness in surpassing current methods for explaining text predictions.


\section*{Acknowledgements}
Most of this work was realized while Damien Garreau was employed at Universit\'e C\^ote d'Azur. 
The authors acknowledge the support of the French Agence Nationale de la Recherche (grant number ANR-21-CE23-0005-01) and EU Horizon 2020 project AI4Media (contract no. 951911). 
The authors are grateful to the OPAL infrastructure from Université Côte d'Azur for providing resources and support. 
Finally, the authors want to thank Sylvain Pogodalla for his pointers regarding word replacements. 


\bibliography{short_biblio}

\begin{thebibliography}{45}
\providecommand{\natexlab}[1]{#1}
\providecommand{\url}[1]{\texttt{#1}}
\expandafter\ifx\csname urlstyle\endcsname\relax
  \providecommand{\doi}[1]{doi: #1}\else
  \providecommand{\doi}{doi: \begingroup \urlstyle{rm}\Url}\fi

\bibitem[Adadi and Berrada(2018)]{adadi2018peeking}
Adadi and Berrada.
\newblock {Peeking inside the black-box: a survey on explainable artificial intelligence}.
\newblock \emph{IEEE access}, 2018.

\bibitem[Amoukou and Brunel(2022)]{amoukou2021consistent}
Amoukou and Brunel.
\newblock {Consistent Sufficient Explanations and Minimal Local Rules for explaining regression and classification models}.
\newblock \emph{{NeurIPS}}, 2022.

\bibitem[Bastings et~al.(2021)]{bastings2021will}
Bastings et~al.
\newblock {"Will You Find These Shortcuts?" A Protocol for Evaluating the Faithfulness of Input Salience Methods for Text Classification}.
\newblock \emph{arXiv:2111.07367}, 2021.

\bibitem[Brown et~al.(2020)]{brown_et_al_2020}
Brown et~al.
\newblock Language models are few-shot learners.
\newblock \emph{{NeurIPS}}, 2020.

\bibitem[Carvalho et~al.(2019)]{carvalho2019machine}
Carvalho et~al.
\newblock Machine learning interpretability: A survey on methods and metrics.
\newblock \emph{Electronics}, 2019.

\bibitem[Chowdhery et~al.(2023)]{chowdhery2023palm}
Chowdhery et~al.
\newblock Palm: Scaling language modeling with pathways.
\newblock \emph{JMLR}, 2023.

\bibitem[Ciravegna et~al.(2021)]{ciravegna2021logic}
Ciravegna et~al.
\newblock Logic explained networks.
\newblock \emph{arXiv:2108.05149}, 2021.

\bibitem[Covert et~al.(2021)]{covert_et_al_2021}
I.~Covert et~al.
\newblock Explaining by removing: A unified framework for model explanation.
\newblock \emph{JMLR}, 2021.

\bibitem[Danilevsky et~al.(2020)]{danilevsky2020survey}
Danilevsky et~al.
\newblock {A Survey of the State of Explainable AI for Natural Language Processing}.
\newblock \emph{AACL-IJCNLP}, 2020.

\bibitem[Delaunay et~al.(2020)]{delaunay2020improving}
Delaunay et~al.
\newblock Improving anchor-based explanations.
\newblock \emph{ACM CIKM}, 2020.

\bibitem[Devlin et~al.(2019)]{devlin2019bert}
Devlin et~al.
\newblock {{BERT}: Pre-training of Deep Bidirectional Transformers for Language Understanding}.
\newblock \emph{NAACL-HLT}, 2019.

\bibitem[DeYoung et~al.(2020)]{deyoung2020eraser}
DeYoung et~al.
\newblock {ERASER: A Benchmark to Evaluate Rationalized NLP Models}.
\newblock \emph{ACL}, 2020.

\bibitem[Garreau and Luxburg(2020)]{garreau2020explaining}
Garreau and Luxburg.
\newblock {Explaining the explainer: A first theoretical analysis of {LIME}}.
\newblock \emph{AISTATS}, 2020.

\bibitem[Guidotti et~al.(2018)]{guidotti2018local}
Guidotti et~al.
\newblock Local rule-based explanations of black box decision systems.
\newblock \emph{arXiv:1805.10820}, 2018.

\bibitem[Hartmann et~al.(2023)]{hartmann2023}
Hartmann et~al.
\newblock More than a feeling: Accuracy and application of sentiment analysis.
\newblock \emph{Int. Journal of Research in Marketing}, 2023.

\bibitem[Hase et~al.(2021)]{hase2021out}
Hase et~al.
\newblock The out-of-distribution problem in explainability and search methods for feature importance explanations.
\newblock \emph{NeurIPS}, 2021.

\bibitem[Kakogeorgiou and Karantzalos(2021)]{kakogeorgiou2021evaluating}
Kakogeorgiou and Karantzalos.
\newblock Evaluating explainable artificial intelligence methods for multi-label deep learning classification tasks in remote sensing.
\newblock \emph{Int. Journal of Applied Earth Observation and Geoinformation}, 2021.

\bibitem[Lakkaraju et~al.(2016)]{lakkaraju2016interpretable}
Lakkaraju et~al.
\newblock Interpretable decision sets: A joint framework for description and prediction.
\newblock \emph{ACM SIGKDD}, 2016.

\bibitem[Lim et~al.(2009)]{lim2009and}
Lim et~al.
\newblock Why and why not explanations improve the intelligibility of context-aware intelligent systems.
\newblock \emph{SIGCHI}, 2009.

\bibitem[Linardatos et~al.(2021)]{linardatos2021explainable}
Linardatos et~al.
\newblock {Explainable AI: A Review of Machine Learning Interpretability Methods}.
\newblock \emph{Entropy}, 2021.

\bibitem[Lipton(2018)]{lipton2018mythos}
Lipton.
\newblock The mythos of model interpretability: In ml, the concept of interpretability is both important and slippery.
\newblock \emph{Queue}, 2018.

\bibitem[Liu et~al.(2019)]{liu2019roberta}
Liu et~al.
\newblock Roberta: A robustly optimized bert pretraining approach.
\newblock \emph{arXiv:1907.11692}, 2019.

\bibitem[Lopardo et~al.(2022)]{lopardo2022smace}
Lopardo et~al.
\newblock {SMACE: A New Method for the Interpretability of Composite Decision Systems}.
\newblock \emph{ECML PKDD}, 2022.

\bibitem[Lopardo et~al.(2023)]{lopardo2022sea}
Lopardo et~al.
\newblock {A Sea of Words: An In-Depth Analysis of Anchors for Text Data}.
\newblock \emph{AISTATS}, 2023.

\bibitem[Luhn(1957)]{luhn1957statistical}
Luhn.
\newblock A statistical approach to mechanized encoding and searching of literary information.
\newblock \emph{IBM Journal of research and development}, 1957.

\bibitem[Lundberg and Lee(2017)]{lundberg2017unified}
Lundberg and Lee.
\newblock {A Unified Approach to Interpreting Model Predictions}.
\newblock \emph{NeurIPS}, 2017.

\bibitem[Malfa et~al.(2021)]{la2021guaranteed}
La~Malfa et~al.
\newblock On guaranteed optimal robust explanations for nlp models.
\newblock \emph{IJCAI}, 2021.

\bibitem[Mardaoui and Garreau(2021)]{mardaoui_garreau_2021}
Mardaoui and Garreau.
\newblock An analysis of {LIME} for text data.
\newblock \emph{AISTATS}, 2021.

\bibitem[Marques-Silva and Ignatiev(2022)]{marques2022delivering}
Marques-Silva and Ignatiev.
\newblock Delivering trustworthy ai through formal xai.
\newblock \emph{AAAI}, 2022.

\bibitem[Minaee et~al.(2024)]{minaee2024large}
Minaee et~al.
\newblock Large language models: A survey.
\newblock \emph{arXiv:2402.06196}, 2024.

\bibitem[Montavon et~al.(2019)]{montavon2019layer}
Montavon et~al.
\newblock Layer-wise relevance propagation: an overview.
\newblock \emph{XAI: interpreting, explaining and visualizing deep learning}, 2019.

\bibitem[Mylonas et~al.(2023)]{mylonas2023attention}
N.~Mylonas et~al.
\newblock An attention matrix for every decision: Faithfulness-based arbitration among multiple attention-based interpretations of transformers in text classification.
\newblock \emph{Data Mining and Knowledge Discovery}, 2023.

\bibitem[Pawelczyk et~al.(2021)]{pawelczyk2021carla}
Pawelczyk et~al.
\newblock Carla: a python library to benchmark algorithmic recourse and counterfactual explanation algorithms.
\newblock \emph{arXiv:2108.00783}, 2021.

\bibitem[Pawelczyk et~al.(2022)]{pawelczyk2022exploring}
Pawelczyk et~al.
\newblock Exploring counterfactual explanations through the lens of adversarial examples: A theoretical and empirical analysis.
\newblock \emph{AISTATS}, 2022.

\bibitem[Ribeiro et~al.(2016)]{ribeiro2016should}
Ribeiro et~al.
\newblock {\say{Why should I trust you?} Explaining the predictions of any classifier}.
\newblock \emph{ACM SIGKDD}, 2016.

\bibitem[Ribeiro et~al.(2018)]{ribeiro2018anchors}
Ribeiro et~al.
\newblock Anchors: High-precision model-agnostic explanations.
\newblock \emph{AAAI}, 2018.

\bibitem[Rigotti et~al.(2021)]{rigotti2021attention}
Rigotti et~al.
\newblock Attention-based interpretability with concept transformers.
\newblock \emph{ICLR}, 2021.

\bibitem[Rockafellar(1997)]{rockafellar1997convex}
Rockafellar.
\newblock \emph{Convex analysis}.
\newblock Princeton university press, 1997.

\bibitem[Sanh et~al.(2019)]{sanh2019distilbert}
Sanh et~al.
\newblock Distilbert, a distilled version of bert: smaller, faster, cheaper and lighter.
\newblock \emph{arXiv:1910.01108}, 2019.

\bibitem[Selvaraju et~al.(2017)]{selvaraju2017grad}
Selvaraju et~al.
\newblock Grad-cam: Visual explanations from deep networks via gradient-based localization.
\newblock \emph{ICCV}, 2017.

\bibitem[Slack et~al.(2020)]{slack2020fooling}
Slack et~al.
\newblock Fooling {LIME} and {SHAP}: Adversarial attacks on post hoc explanation methods.
\newblock \emph{AAAI/ACM}, 2020.

\bibitem[Stumpf et~al.(2007)]{stumpf2007toward}
Stumpf et~al.
\newblock Toward harnessing user feedback for machine learning.
\newblock \emph{ACM IUI}, 2007.

\bibitem[Touvron et~al.(2023)]{touvron2023llama}
Touvron et~al.
\newblock Llama: Open and efficient foundation language models.
\newblock \emph{arXiv:2302.13971}, 2023.

\bibitem[Wachter et~al.(2017)]{wachter2017counterfactual}
Wachter et~al.
\newblock Counterfactual explanations without opening the black box: Automated decisions and the gdpr.
\newblock \emph{Harv. JL \& Tech.}, 2017.

\bibitem[Wang and Rudin(2015)]{wang2015falling}
Wang and Rudin.
\newblock Falling rule lists.
\newblock \emph{AISTATS}, 2015.

\end{thebibliography}

\newpage
\appendix

\begin{center}
{\Large Appendix for the paper \\
\say{Faithful and Local Interpretability for Textual Predictions}}
\end{center}

\paragraph{Organization of the Appendix.}
In Section~\ref{sec:proofs}, we provide theoretical proofs for the results presented in the paper. 
In Section~\ref{appendix-sec:experiments}, we expose implementation details and provide additional results. 
For any other experimental detail, we released the code for FRED and for experiments at \url{https://github.com/gianluigilopardo/fred}. 

\section{Proofs}
\label{sec:proofs}
In this section, we collect the proofs of the theoretical results presented in the main paper. 
First, we prove Lemma~\ref{lemma:empirical-drop} and Lemma~\ref{lemma:n-sample} from Section~\ref{sec:FRED}. 
Then, we move to the mathematical validation of our analysis presented in Section~\ref{sec:analysis} by proving Proposition~\ref{prop:linear-models} and Proposition~\ref{prop:presence-words}. 
\subsection{\texorpdfstring{Proof of Lemma \ref{lemma:empirical-drop}: Convergence of Empirical Drop $\Empdrop_\candidate$}{}}
\label{proof-lemma-empirical-drop}
%
Let $n$ denote the sample size. 
We are interested in the empirical average of predictions on instances that do not contain the candidate $\candidate$, \emph{i.e.}, $\frac{1}{n_\candidate} \sum_{\candidate \notin x_i} \model(x_i)$. 
This quantity can be rewritten using indicator functions, as
\[ 
\frac{1}{n_\candidate} \sum_{\candidate \notin x_i} \model(x_i) = \frac{n}{n_\candidate} \frac{1}{n}\sum_{i=1}^n \model(x_i)\indic{\candidate \notin x_i} \, .
\]
We can therefore apply the weak law of large numbers. 
As $n$ grows infinitely, this empirical average converges in probability to the expected value of the predictions for instances without $\candidate$:
\[ 
\frac{1}{n}\sum_{i=1}^n \model(x_i)\indic{\candidate \notin x_i} \cvproba \expec{\model(x) \indic{\candidate \notin x}} 
\,.\]

Likewise, as $n\to + \infty$, 
$\frac{n}{n_\candidate}$ converges in probability to $\frac{1}{\proba{\candidate \notin x}}$. 
By multiplying the two limits, we get:
\[ \frac{1}{n_\candidate} \sum_{\candidate \notin x_i} \model(x_i) \cvproba \frac{\expec{\model(x) \indic{\candidate \notin x}}}{\proba{\candidate \notin x}} \,.\]

This approximation holds true when both $n$ and $n_\candidate$ are sufficiently large, facilitating the estimation of the drop in prediction associated to a candidate. 
\qed

\subsection{\texorpdfstring{Proof of Lemma \ref{lemma:n-sample}: Choosing $n$}{}}
\label{proof-lemma-n-sample}
For any candidate $\candidate$ with size $\length = \size{\candidate}$, and any sample $x_i$, $i \in [n]$, $\proba{\candidate \notin x_i} = \frac{1}{2^\length}$, since word removals are \emph{i.i.d.} with probability $1/2$. 
Then, 
\begin{align*}
    \proba{\exists i \in [n] : \candidate \notin x_i} & = \sum_{k=1}^n \binom{n}{k} \left(\frac{1}{2^\length}\right)^k \left(1 - \frac{1}{2^\length}\right)^{n-k} \\
    & = 1 - \left(1 - 1/2^\length\right)^n \\
    \proba{\exists i \in [n] : \candidate \notin x_i} & \geq \alpha \iff n \geq \frac{\logs{1 - \alpha}}{\logs{1 - 1/2^\length}}
    \,. 
\end{align*}
Since a candidate has maximal length $\length_{max}$, we can choose the sample size by setting $n = \ceil{\frac{\logs{1 - \alpha}}{\logs{1 - 1/2^{\length_{max}}}}}$. 
\qed

\subsection{Proof of Proposition \ref{prop:linear-models}: Linear models}
\label{proof-prop-linear-models}
%
Let $\candrop_\candidate$ be the drop in prediction associated to candidate $\candidate$. 
Let $\lambda,\lambda_0$ be the coefficients associated to the linear classifier $\model$ defined by Eq.~\eqref{eq:def-linear-classifier}. 
Assume $\lambda_1 \idf_1 > \lambda_2 \idf_2 > \cdots > \lambda_d \idf_d$. 
Then, 
\begin{align*}
    \candrop_\candidate & = \expecunder{\drop(x)}{c} = \expec{\model(x)} - \expecunder{\model(x)}{c} = \expec{\lambda^\top\tfidf{\xi}} - \expecunder{\lambda^\top\tfidf{x})}{c} \\
    & = \expec{\sum_{j=1}^d \lambda_j\idf_j\Mult_j} - \expecunder{\sum_{j=1}^d \lambda_j\idf_j\Mult_j}{c} \\
    & = \sum_{j=1}^d \lambda_j\idf_j\expec{\Mult_j} - \sum_{j=1}^d \lambda_j\idf_j\expecunder{\Mult_j}{\candidate_j}    \tag{since removals are \emph{i.i.d.}} \\    
    & = \sum_{j=1}^d \lambda_j\idf_j p \mult_j - \left(\sum_{\candidate_j = 0} \lambda_j\idf_j\expec{\Mult_j} + \sum_{\candidate_i \neq 0} \lambda_i\idf_i\expec{\Mult_i} \right) \\ 
    & = \sum_{j=1}^d \lambda_j\idf_j p \mult_j - \left(\sum_{\candidate_j = 0} \lambda_j\idf_j p\mult_j + \sum_{\candidate_i \neq 0} \lambda_i\idf_i p(\mult_i-1) \right)     \tag{since $\Mult_i \sim B(\mult_i-\candidate_i, p)$} \\ 
    & = p\sum_{\candidate_i \neq 0}\lambda_i\idf_i \candidate_i, 
    \,,
\end{align*}
and, for any fixed length $\length$, this quantity is maximized by the subset $\candidate$ having $\size{\candidate} = \length$ and containing the firsts $\length$ words ranked by $\lambda_j\idf_j$.     
\qed

\subsection{Proof of Proposition \ref{prop:presence-words}: Presence of shortcuts}
\label{proof-prop-presence-words}
%
Let us assume $\model(\xi) = 1$ (the case $\model(\xi) = 0$ is specular). 
We can rewrite the drop in prediction associated to the candidate $\candidate$ as follows. 
\begin{align*}
    \candrop_\candidate & = \expecunder{\drop(x)}{\candidate} = \expec{\model(x)} - \expecunder{\model(x)}{\candidate} = \expec{\prod_{j \in J} \indic{\word_j \in x}} - \expecunder{\prod_{j \in J} \indic{\word_j \in x}}{\candidate} \\
    & = \expec{\prod_{j \in J} \indic{\tfidf{x}_j > 0}} - \expecunder{\prod_{j \in J} \indic{\tfidf{x}_j > 0}}{\candidate} \\
    & = \prod_{j \in J}\expec{\indic{\tfidf{x}_j > 0}} - \prod_{j \in J}\expecunder{\indic{\tfidf{x}_j > 0}}{\candidate}       \tag{since removals are \emph{i.i.d.}} \\
    & = \prod_{j \in J}\proba{\tfidf{x}_j > 0} - \prod_{j \in J}\probaunder{\tfidf{x}_j > 0}{\candidate} \\
    & = \prod_{j \in J}\proba{\Mult_j\idf_j > 0} - \prod_{j \in J}\probaunder{\Mult_j\idf_j > 0}{\candidate}     \tag{since $\tfidf{x}_j = \Mult_j\idf_j$} \\
    & = \prod_{j \in J}\proba{\Mult_j > 0} - \prod_{j \in J}\probaunder{\Mult_j > 0}{\candidate}       \tag{since $\idf_j > 0$ for all $j \in [d]$} \\
    & = \text{const} - \prod_{j \in J}(1 -\probaunder{\Mult_j = 0}{\candidate}) \\
    & = \text{const} - \prod_{j \in J}(1 - (1-p)^{\mult_j - \candidate_j}) \\
    & = \text{const} - \prod_{j \in J}(1 - (1-p)^{\mult_j - \candidate_j}) \,. \\
\end{align*}

Thus, the following conditions are equivalent. 
\begin{align}
    \Maximize \quad \candrop_\candidate & \iff
    \Maximize \quad \text{const} - \prod_{j \in J}(1 - (1-p)^{\mult_j - \candidate_j})     \nonumber \\
    & \iff \Minimize \quad \prod_{j \in J}(1 - p^{\mult_j - \candidate_j})     \tag{we use $p=1/2$} \nonumber \\
    & \iff \Minimize \quad \prod_{j \in J}(1 - p^{\mult_j - \candidate_j})   \label{eq:rule-minimization} \\
    & \iff \Minimize \quad \logs{\prod_{j \in J}(1 - p^{\mult_j - \candidate_j})}     \nonumber \\
    & \iff \Minimize \quad \sum_{j \in J}\logs{1 - p^{\mult_j - \candidate_j}}     \label{eq:rule-minimization-log}
    \,. 
\end{align}
Let us first consider Eq. \eqref{eq:rule-minimization}.  
We study the problem for a length $\length \leq \length_{max}$ of the candidates, \emph{i.e.}, the problem is 
\begin{align}
\label{prob:rule-minimization}
    \Minimize \quad           & F(c) \defeq \prod_{j \in J}(1 - p^{\mult_j - \candidate_j}) \\
    \text{subject to} \quad   & \sum_{j=1}^d \candidate_j = \length     \nonumber \\
    \text{and} \quad          & \candidate_j \in [\mult_j] \,, \quad j \in [d]       \nonumber \,.
\end{align}
$F(c) = 0$ is the global minimum. 
We split the proof in three cases: (1) $\length = \mult_1$, (2) $\length > \mult_1$, and (3) $\length < \mult_1$. 

\paragraph{(1) $\length = \mult_1$.} 
When $\length$ is equal to the smallest multiplicity, the optimal candidate $\candidate^\star$ is such that $\candidate^\star_1 = \mult_1$ and $\candidate^\star_j = 0$ for $j \neq 1$. 
Indeed, $F(\candidate^\star) = 0 < F(\candidate)$ for any $\candidate \neq \candidate^\star$ such that $\size{\candidate} = \size{\candidate^\star}$. 

\paragraph{(2) $\length > \mult_1$.}
The previous paragraph implies that the optimal candidate $\candidate^\star$ is always such that $\size{\candidate^\star} \leq \mult_1$. 
Indeed, any candidate ${\candidate}$ of size $\length$ such that $F({\candidate}) = F(\candidate^\star) = 0$ has size $\size{{\candidate}} > \mult_1 \geq \size{\candidate^\star}$. 
Hence, we can disregard this case. %

\paragraph{(3) $\length < \mult_1$.}
The optimal candidate is $\candidate^\star$ with $\candidate^\star_1 = \length$ and $\candidate^\star_j = 0$ for $j \neq 1$. 
To prove this, let us study the continuous problem from Eq. \eqref{eq:rule-minimization-log}: 

\begin{align}
\label{prob:rule-minimization-log}
    \Minimize \quad           & F(c) \defeq \sum_{j \in J}\logs{1 - p^{\mult_j - x_j}} \\
    \text{subject to} \quad   & \sum_{j=1}^d x_j = \length     \nonumber \\
    \text{and} \quad          & 0 \leq x_j \leq \mult_j \,, \quad j \in [d]       \nonumber 
    \,.
\end{align}

First, we notice that Problem \eqref{prob:rule-minimization-log} consists of minimizing a concave function over a convex set, indeed, since the problem is separable the Hessian matrix $H$ of $F$ is diagonal and defined, for $i,j\in[d]$, as 
\[
(H)_{i,j} = \begin{cases}
    - \frac{(\logs{p})^2 p^{\mult_j - x}}{(p^x - p^{\mult_j})^2} < 0 \,, \qquad \text{if } i = j \,, \\
    0 \,, \qquad \text{if } i \neq j 
    \, .
\end{cases}
\]

As a consequence, the solutions are found on the corners (see Corollary~32.3.1, \citet{rockafellar1997convex}), defined as $(\length, 0, \ldots, 0)$, $(0, \length, \ldots, 0)$, $\ldots$, $(0, 0, \ldots, \length)$. 
Finally, we notice that the candidate $\candidate$ minimizing Eq.~\eqref{prob:rule-minimization-log} is such that $\mult_j - \candidate_j = \mult_j - \length$ is minima, \emph{i.e.}, $\candidate_1 = \length$ and $\candidate_i = 0$ for $i \in [d]$. 
\qed
%

%
%
\section{Experiments}
\label{appendix-sec:experiments}
In this section, we report details of experiments and additional results omitted in the main paper due to space constraints. 

\subsection{Setting}
\label{sec:app-setting}
All the experiments reported in this Section and in the main paper are implemented in \texttt{Python} and ran on GPU \texttt{Nvidia A100}. 
The code for FRED and the experiments is available at \url{https://github.com/gianluigilopardo/fred}. 

\paragraph{Datasets.} 
We employ three sentiment analysis datasets: Restaurants\footnote{https://www.kaggle.com/hj5992/restaurantreviews}, Yelp reviews\footnote{https://www.kaggle.com/omkarsabnis/yelp-reviews-dataset}, and IMDb\footnote{https://huggingface.co/datasets/imdb}, and the Tweets hate speech detection\footnote{https://huggingface.co/datasets/tweets\_hate\_speech\_detection} dataset. 
Note that the datasets have a substantial difference in document length: while the average length of tokens in Restaurants and Tweets is about $10$, for Yelp it is about $150$, in IMDb it is $230$. 
As shown below, this difference has a relevant impact on explainers' behavior. 

\paragraph{Models.} 
We trained a logistic classifier, a decision tree, and a random forest classifier on each dataset (with the default parameters of \url{https://scikit-learn.org/}). 
Additionally, we employed a pretrained version of RoBERTa\footnote{https://huggingface.co/siebert/sentiment-roberta-large-english} \citep{hartmann2023} and DistilBERT\footnote{https://huggingface.co/distilbert/distilbert-base-uncased} for the three sentiment analysis datasets. 
These are two cutting-edge transformer models for sentiment analysis sourced from the Hugging Face repository.
Finally, we remark that we always consider documents with positive predictions, and we explain the positive class. 

Table \ref{tab:model_accuracy} reports the accuracies for each model and dataset. 
Remark that logistic classifiers, decision trees, random forest classifiers have been trained on each dataset according to their task. 
Contrarily, we use pre-trained version of Roberta and DistilBERT on the datasets. 
\begin{table}[t]
\centering
\caption{\label{tab:model_accuracy}Accuracy of machine learning models evaluated on datasets used in the experiments. 
An asterisk (*) denotes models that were trained specifically on the corresponding dataset. 
RoBERTa and DistilBERT are leveraged here as pre-trained models, meaning their weights were not further fine-tuned on the individual datasets. 
}
\begin{tabular}{ccc}
\hline
Dataset & Model & Accuracy \\
\hline
\multirow{5}{*}{Restaurants} 
& Logistic Regression* & $0.808$  \\ 
& Decision Tree* & $0.676$ \\ 
& Random Forest* & $0.748$  \\ 
& RoBERTa & $0.972$ \\ 
& DistilBERT & $0.500$ \\  
\hline
\multirow{5}{*}{Yelp} 
& Logistic Regression*& $0.892$ \\ 
& Decision Tree* & $0.748$\\ 
& Random Forest* & $0.868$ \\ 
& RoBERTa & $0.988$ \\ 
& DistilBERT & $0.472$ \\  
\hline
\multirow{5}{*}{IMDB} 
& Logistic Regression*& $0.710$  \\ 
& Decision Tree* & $0.700$ \\ 
& Random Forest* & $0.834$  \\ 
& RoBERTa & $0.950$  \\ 
& DistilBERT & $0.500$ \\  
\hline
\multirow{5}{*}{Tweets} 
& Logistic Regression*& $0.930$ \\ 
& Decision Tree* & $0.941$ \\ 
& Random Forest* & $0.959$  \\ 
& RoBERTa & $0.273$  \\ 
& DistilBERT & $0.929$  \\  
\hline
\end{tabular}
\end{table}

\paragraph{Explainers.}
FRED is compared to three well-known explainers: Anchors, SHAP, and LIME. 
We use the official implementation (respectively available and licensed at \url{https://github.com/marcotcr/anchor}, \url{https://github.com/shap/shap}, \url{https://github.com/marcotcr/lime}) of these methods with all default parameters. 

Remark that while LIME and SHAP share the principle of attributing an importance score for each token, Anchor aim at outputting the most significant subset of tokens (while FRED has both the options). 
We apply the following scheme to perform a fair evaluation. 
We use as explanation set $\explanation$ for LIME and SHAP the first $k$ tokens ranked by importance, where $k$ is the size of FRED's explanation (\texttt{fredpos}) for the same example. 
This is used for computing comprehensiveness, sufficiency, robustness, and proportion. 
Conversely, for computing the AUC-MoRF of Anchors we select the tokens ordered as in the anchor. 

Finally, the metrics are computed over the first $100$ instances ranked by length of positively classified documents from each test set. 

\subsection{Additional experimental results}
\label{sec:app-additional}
We report below additional experimental results omitted in the main paper due to space constraints. 
Again, both FRED-mask and FRED-pos produce more faithful explanations than the others. 
In general, Anchor performs well on small documents (Restaurant dataset), while its behavior on larger documents is highly nonlinear, as it tends to be conservative on the size of the anchor. 
LIME and SHAP have a close behavior, as somehow expected, but the latter is significantly more efficient. 
On small documents, SHAP performs an exhaustive search over all possible token subsets, motivating its high robustness. 
\begin{table}[ht]
\caption{\label{exp-tab:restaurants_logistic_1}
Comparison on a logistic classifier for Restaurant reviews ($p=0.1$, $\epsilon=0.15$).}
\centering
\setlength{\tabcolsep}{2pt}
\begin{tabular}{@{}l|rrrrrrr@{}}
\textbf{} & \texttt{suffic.} $\downarrow$ & \texttt{compreh.} $\uparrow$ & \texttt{robust}. $\uparrow$ & \texttt{aucmorf} $\downarrow$ & \texttt{time (s)} $\downarrow$ & \texttt{proport.} $\downarrow$ & \\
\hline
fred & $-0.118 (0.09)$ & $0.130 (0.05)$ & $0.866 (0.20)$ & $0.709 (0.09)$ & $0.055 (0.02)$ & $0.227 (0.13)$ & \\
fredpos & $-0.094 (0.11)$ & $0.097 (0.06)$ & $0.876 (0.21)$ & $0.721 (0.09)$ & $0.059 (0.02)$ & $0.159 (0.09)$ & \\
lime & $-0.111 (0.09)$ & $0.102 (0.06)$ & $0.968 (0.10)$ & $0.726 (0.09)$ & $0.130 (0.02)$ & $0.159 (0.09)$ & \\
shap & $-0.120 (0.09)$ & $0.100 (0.06)$ & $1.000 (0.00)$ & $0.712 (0.09)$ & $0.065 (0.37)$ & $0.159 (0.09)$ & \\
anchor & $-0.058 (0.15)$ & $0.072 (0.06)$ & $0.772 (0.30)$ & $0.874 (0.10)$ & $0.121 (0.19)$ & $0.136 (0.07)$ & \\
\end{tabular}
\end{table}

\begin{table}[ht]
\caption{\label{exp-tab:restaurants_logistic_5}
Comparison on a logistic classifier for Restaurant reviews ($p=0.5$, $\epsilon=0.15$).}
\centering
\setlength{\tabcolsep}{2pt}
\begin{tabular}{@{}l|rrrrrrr@{}}
\textbf{} & \texttt{suffic.} $\downarrow$ & \texttt{compreh.} $\uparrow$ & \texttt{robust}. $\uparrow$ & \texttt{aucmorf} $\downarrow$ & \texttt{time (s)} $\downarrow$ & \texttt{proport.} $\downarrow$ & \\
\hline
fred & $-0.119 (0.09)$ & $0.120 (0.05)$ & $0.926 (0.15)$ & $0.707 (0.09)$ & $0.075 (0.03)$ & $0.216 (0.14)$ & \\
fredpos & $-0.089 (0.11)$ & $0.086 (0.07)$ & $0.924 (0.20)$ & $0.718 (0.09)$ & $0.075 (0.02)$ & $0.128 (0.07)$ & \\
lime & $-0.105 (0.10)$ & $0.093 (0.06)$ & $0.982 (0.09)$ & $0.726 (0.09)$ & $0.128 (0.02)$ & $0.128 (0.07)$ & \\
shap & $-0.114 (0.09)$ & $0.092 (0.06)$ & $1.000 (0.00)$ & $0.712 (0.09)$ & $0.065 (0.36)$ & $0.128 (0.07)$ & \\
anchor & $-0.054 (0.15)$ & $0.069 (0.07)$ & $0.784 (0.30)$ & $0.874 (0.10)$ & $0.122 (0.19)$ & $0.127 (0.07)$ & \\
\end{tabular}
\end{table}

\begin{table}[ht]
\caption{\label{exp-tab:restaurants_tree_1}
Comparison on a decision tree for Restaurant reviews ($p=0.1$, $\epsilon=0.15$).}
\centering
\setlength{\tabcolsep}{2pt}
\begin{tabular}{@{}l|rrrrrrr@{}}
\textbf{} & \texttt{suffic.} $\downarrow$ & \texttt{compreh.} $\uparrow$ & \texttt{robust}. $\uparrow$ & \texttt{aucmorf} $\downarrow$ & \texttt{time (s)} $\downarrow$ & \texttt{proport.} $\downarrow$ & \\
\hline
fred & $0.030 (0.17)$ & $0.940 (0.24)$ & $0.960 (0.14)$ & $0.138 (0.14)$ & $0.057 (0.02)$ & $0.158 (0.09)$ & \\
fredpos & $0.040 (0.20)$ & $0.750 (0.43)$ & $0.947 (0.15)$ & $0.183 (0.16)$ & $0.062 (0.02)$ & $0.137 (0.08)$ & \\
lime & $0.020 (0.14)$ & $0.750 (0.43)$ & $0.894 (0.21)$ & $0.110 (0.09)$ & $0.132 (0.01)$ & $0.137 (0.08)$ & \\
shap & $0.010 (0.10)$ & $0.690 (0.46)$ & $1.000 (0.00)$ & $0.132 (0.11)$ & $0.054 (0.20)$ & $0.137 (0.08)$ & \\
anchor & $0.020 (0.14)$ & $0.630 (0.48)$ & $0.968 (0.15)$ & $0.381 (0.45)$ & $0.190 (0.76)$ & $0.111 (0.05)$ & \\
\end{tabular}
\end{table}

\begin{table}[ht]
\caption{\label{exp-tab:restaurants_tree_5}
Comparison on a decision tree for Restaurant reviews ($p=0.5$, $\epsilon=0.15$).}
\centering
\setlength{\tabcolsep}{2pt}
\begin{tabular}{@{}l|rrrrrrr@{}}
\textbf{} & \texttt{suffic.} $\downarrow$ & \texttt{compreh.} $\uparrow$ & \texttt{robust}. $\uparrow$ & \texttt{aucmorf} $\downarrow$ & \texttt{time (s)} $\downarrow$ & \texttt{proport.} $\downarrow$ & \\
\hline
fred & $0.020 (0.14)$ & $0.620 (0.49)$ & $0.807 (0.28)$ & $0.116 (0.09)$ & $0.067 (0.02)$ & $0.116 (0.05)$ & \\
fredpos & $0.030 (0.17)$ & $0.650 (0.48)$ & $0.938 (0.17)$ & $0.167 (0.17)$ & $0.080 (0.03)$ & $0.113 (0.05)$ & \\
lime & $0.020 (0.14)$ & $0.650 (0.48)$ & $0.822 (0.27)$ & $0.110 (0.09)$ & $0.132 (0.01)$ & $0.113 (0.05)$ & \\
shap & $0.010 (0.10)$ & $0.600 (0.49)$ & $1.000 (0.00)$ & $0.132 (0.11)$ & $0.051 (0.15)$ & $0.113 (0.05)$ & \\
anchor & $0.020 (0.14)$ & $0.630 (0.48)$ & $0.968 (0.15)$ & $0.381 (0.45)$ & $0.190 (0.76)$ & $0.111 (0.05)$ & \\
\end{tabular}
\end{table}

\begin{table}[ht]
\caption{\label{exp-tab:restaurants_forest_1}
Comparison on a random forest classifier for Restaurant reviews ($p=0.1$, $\epsilon=0.15$).}
\centering
\setlength{\tabcolsep}{2pt}
\begin{tabular}{@{}l|rrrrrrr@{}}
\textbf{} & \texttt{suffic.} $\downarrow$ & \texttt{compreh.} $\uparrow$ & \texttt{robust}. $\uparrow$ & \texttt{aucmorf} $\downarrow$ & \texttt{time (s)} $\downarrow$ & \texttt{proport.} $\downarrow$ & \\
\hline
fred & $0.040 (0.15)$ & $0.192 (0.09)$ & $0.919 (0.17)$ & $0.562 (0.13)$ & $0.134 (0.02)$ & $0.160 (0.07)$ & \\
fredpos & $0.084 (0.17)$ & $0.156 (0.11)$ & $0.942 (0.18)$ & $0.544 (0.11)$ & $0.140 (0.02)$ & $0.126 (0.06)$ & \\
lime & $0.061 (0.14)$ & $0.157 (0.11)$ & $0.974 (0.12)$ & $0.516 (0.10)$ & $0.218 (0.01)$ & $0.126 (0.06)$ & \\
shap & $0.054 (0.13)$ & $0.150 (0.11)$ & $1.000 (0.00)$ & $0.495 (0.09)$ & $0.172 (0.21)$ & $0.126 (0.06)$ & \\
anchor & $0.069 (0.15)$ & $0.129 (0.13)$ & $0.877 (0.24)$ & $0.778 (0.18)$ & $0.733 (0.88)$ & $0.118 (0.05)$ & \\
\end{tabular}
\end{table}

\begin{table}[ht]
\caption{\label{exp-tab:restaurants_forest_5}
Comparison on a random forest classifier for Restaurant reviews ($p=0.5$, $\epsilon=0.15$).}
\centering
\setlength{\tabcolsep}{2pt}
\begin{tabular}{@{}l|rrrrrrr@{}}
\textbf{} & \texttt{suffic.} $\downarrow$ & \texttt{compreh.} $\uparrow$ & \texttt{robust}. $\uparrow$ & \texttt{aucmorf} $\downarrow$ & \texttt{time (s)} $\downarrow$ & \texttt{proport.} $\downarrow$ & \\
\hline
fred & $0.047 (0.15)$ & $0.162 (0.12)$ & $0.955 (0.15)$ & $0.512 (0.11)$ & $0.219 (0.03)$ & $0.146 (0.08)$ & \\
fredpos & $0.072 (0.15)$ & $0.156 (0.12)$ & $0.942 (0.18)$ & $0.523 (0.10)$ & $0.233 (0.03)$ & $0.138 (0.08)$ & \\
lime & $0.063 (0.15)$ & $0.162 (0.12)$ & $0.980 (0.09)$ & $0.516 (0.09)$ & $0.329 (0.02)$ & $0.138 (0.08)$ & \\
shap & $0.055 (0.14)$ & $0.157 (0.12)$ & $1.000 (0.00)$ & $0.500 (0.09)$ & $0.187 (0.27)$ & $0.138 (0.08)$ & \\
anchor & $0.075 (0.15)$ & $0.139 (0.13)$ & $0.862 (0.27)$ & $0.770 (0.17)$ & $0.837 (0.98)$ & $0.135 (0.08)$ & \\
\end{tabular}
\end{table}

\begin{table}[ht]
\caption{\label{exp-tab:restaurants_distilbert_1}
Comparison on DistilBERT for Restaurant reviews ($p=0.1$, $\epsilon=0.15$).}
\centering
\setlength{\tabcolsep}{2pt}
\begin{tabular}{@{}l|rrrrrrr@{}}
\textbf{} & \texttt{suffic.} $\downarrow$ & \texttt{compreh.} $\uparrow$ & \texttt{robust}. $\uparrow$ & \texttt{aucmorf} $\downarrow$ & \texttt{time (s)} $\downarrow$ & \texttt{proport.} $\downarrow$ & \\
\hline
fred & $-0.017 (0.01)$ & $0.003 (0.00)$ & $0.910 (0.17)$ & $0.997 (0.01)$ & $7.273 (0.40)$ & $0.145 (0.17)$ & \\
fredpos & $-0.014 (0.01)$ & $-0.004 (0.01)$ & $0.571 (0.26)$ & $1.002 (0.01)$ & $7.271 (0.32)$ & $0.363 (0.23)$ & \\
lime & $-0.018 (0.01)$ & $0.001 (0.00)$ & $0.943 (0.17)$ & $0.998 (0.01)$ & $7.806 (0.34)$ & $0.164 (0.16)$ & \\
shap & $-0.014 (0.01)$ & $-0.004 (0.01)$ & $1.000 (0.00)$ & $1.015 (0.01)$ & $0.183 (0.20)$ & $0.363 (0.23)$ & \\
anchor & $-0.016 (0.01)$ & $-0.003 (0.00)$ & $1.000 (0.00)$ & $1.008 (0.01)$ & $0.500 (0.23)$ & $0.121 (0.07)$ & \\
\end{tabular}
\end{table}

\begin{table}[ht]
\caption{\label{exp-tab:restaurants_distilbert_5}
Comparison on DistilBERT for Restaurant reviews ($p=0.5$, $\epsilon=0.15$).}
\centering
\setlength{\tabcolsep}{2pt}
\begin{tabular}{@{}l|rrrrrrr@{}}
\textbf{} & \texttt{suffic.} $\downarrow$ & \texttt{compreh.} $\uparrow$ & \texttt{robust}. $\uparrow$ & \texttt{aucmorf} $\downarrow$ & \texttt{time (s)} $\downarrow$ & \texttt{proport.} $\downarrow$ & \\
\hline
fred & $0.003 (0.01)$ & $0.023 (0.01)$ & $0.961 (0.08)$ & $0.970 (0.01)$ & $7.331 (0.40)$ & $0.700 (0.18)$ & \\
fredpos & $0.007 (0.01)$ & $0.015 (0.01)$ & $0.882 (0.15)$ & $0.978 (0.02)$ & $7.372 (0.30)$ & $0.461 (0.32)$ & \\
lime & $0.006 (0.01)$ & $0.017 (0.01)$ & $0.982 (0.08)$ & $0.971 (0.01)$ & $7.833 (0.53)$ & $0.455 (0.31)$ & \\
shap & $0.007 (0.01)$ & $0.015 (0.01)$ & $1.000 (0.00)$ & $0.974 (0.01)$ & $0.180 (0.20)$ & $0.461 (0.32)$ & \\
anchor & $0.014 (0.01)$ & $0.006 (0.01)$ & $0.912 (0.25)$ & $0.988 (0.01)$ & $0.981 (1.31)$ & $0.130 (0.09)$ & \\
\end{tabular}
\end{table}

\begin{table}[ht]
\caption{\label{exp-tab:restaurants_roberta_1}
Comparison on Roberta for Restaurant reviews ($p=0.1$, $\epsilon=0.15$).}
\centering
\setlength{\tabcolsep}{2pt}
\begin{tabular}{@{}l|rrrrrrr@{}}
\textbf{} & \texttt{suffic.} $\downarrow$ & \texttt{compreh.} $\uparrow$ & \texttt{robust}. $\uparrow$ & \texttt{aucmorf} $\downarrow$ & \texttt{time (s)} $\downarrow$ & \texttt{proport.} $\downarrow$ & \\
\hline
fred & $0.487 (0.49)$ & $0.766 (0.42)$ & $0.916 (0.20)$ & $0.151 (0.12)$ & $35.276 (1.48)$ & $0.158 (0.08)$ & \\
fredpos & $0.577 (0.49)$ & $0.528 (0.50)$ & $0.953 (0.18)$ & $0.230 (0.18)$ & $34.953 (1.49)$ & $0.130 (0.06)$ & \\
lime & $0.507 (0.49)$ & $0.538 (0.50)$ & $0.930 (0.21)$ & $0.162 (0.13)$ & $38.554 (2.36)$ & $0.130 (0.06)$ & \\
shap & $0.756 (0.42)$ & $0.323 (0.46)$ & $1.000 (0.00)$ & $0.306 (0.21)$ & $1.814 (1.58)$ & $0.130 (0.06)$ & \\
anchor & $0.509 (0.49)$ & $0.538 (0.50)$ & $0.850 (0.32)$ & $0.454 (0.45)$ & $15.556 (20.26)$ & $0.126 (0.06)$ & \\
\end{tabular}
\end{table}

\begin{table}[ht]
\caption{\label{exp-tab:yelp_logistic_1}
Comparison on logistic classifier for Yelp reviews ($p=0.1$, $\epsilon=0.15$).}
\centering
\setlength{\tabcolsep}{2pt}
\begin{tabular}{@{}l|rrrrrrr@{}}
\textbf{} & \texttt{suffic.} $\downarrow$ & \texttt{compreh.} $\uparrow$ & \texttt{robust}. $\uparrow$ & \texttt{aucmorf} $\downarrow$ & \texttt{time (s)} $\downarrow$ & \texttt{proport.} $\downarrow$ & \\
\hline
fred & $-0.200 (0.09)$ & $0.101 (0.03)$ & $0.551 (0.27)$ & $0.722 (0.07)$ & $0.251 (0.15)$ & $0.088 (0.07)$ & \\
fredpos & $-0.193 (0.09)$ & $0.088 (0.03)$ & $0.571 (0.29)$ & $0.721 (0.07)$ & $0.266 (0.16)$ & $0.075 (0.06)$ & \\
lime & $-0.207 (0.09)$ & $0.094 (0.03)$ & $0.897 (0.18)$ & $0.790 (0.07)$ & $0.280 (0.12)$ & $0.075 (0.06)$ & \\
shap & $-0.209 (0.09)$ & $0.086 (0.03)$ & $1.000 (0.00)$ & $0.748 (0.07)$ & $0.188 (0.49)$ & $0.075 (0.06)$ & \\
anchor & $-0.012 (0.23)$ & $0.034 (0.05)$ & $0.658 (0.37)$ & $0.948 (0.07)$ & $0.983 (2.53)$ & $0.041 (0.05)$ & \\
\end{tabular}
\end{table}

\begin{table}[ht]
\caption{\label{exp-tab:yelp_logistic_5}
Comparison on logistic classifier for Yelp reviews ($p=0.5$, $\epsilon=0.15$).}
\centering
\setlength{\tabcolsep}{2pt}
\begin{tabular}{@{}l|rrrrrrr@{}}
\textbf{} & \texttt{suffic.} $\downarrow$ & \texttt{compreh.} $\uparrow$ & \texttt{robust}. $\uparrow$ & \texttt{aucmorf} $\downarrow$ & \texttt{time (s)} $\downarrow$ & \texttt{proport.} $\downarrow$ & \\
\hline
fred & $-0.205 (0.09)$ & $0.103 (0.02)$ & $0.749 (0.23)$ & $0.724 (0.07)$ & $0.355 (0.21)$ & $0.090 (0.08)$ & \\
fredpos & $-0.188 (0.10)$ & $0.068 (0.04)$ & $0.754 (0.29)$ & $0.718 (0.07)$ & $0.398 (0.24)$ & $0.069 (0.09)$ & \\
lime & $-0.196 (0.10)$ & $0.067 (0.04)$ & $0.866 (0.23)$ & $0.790 (0.07)$ & $0.287 (0.12)$ & $0.060 (0.07)$ & \\
shap & $-0.188 (0.10)$ & $0.066 (0.04)$ & $1.000 (0.00)$ & $0.747 (0.08)$ & $0.171 (0.30)$ & $0.069 (0.09)$ & \\
anchor & $-0.031 (0.23)$ & $0.037 (0.05)$ & $0.629 (0.39)$ & $0.945 (0.07)$ & $1.123 (2.85)$ & $0.043 (0.06)$ & \\
\end{tabular}
\end{table}

\begin{table}[ht]
\caption{\label{exp-tab:yelp_tree_1}
Comparison on decision tree for Yelp reviews ($p=0.1$, $\epsilon=0.15$).}
\centering
\setlength{\tabcolsep}{2pt}
\begin{tabular}{@{}l|rrrrrrr@{}}
\textbf{} & \texttt{suffic.} $\downarrow$ & \texttt{compreh.} $\uparrow$ & \texttt{robust}. $\uparrow$ & \texttt{aucmorf} $\downarrow$ & \texttt{time (s)} $\downarrow$ & \texttt{proport.} $\downarrow$ & \\
\hline
fred & $0.240 (0.43)$ & $0.840 (0.37)$ & $0.899 (0.23)$ & $0.144 (0.27)$ & $0.292 (0.17)$ & $0.042 (0.05)$ & \\
fredpos & $0.250 (0.43)$ & $0.810 (0.39)$ & $0.842 (0.31)$ & $0.194 (0.33)$ & $0.320 (0.18)$ & $0.041 (0.05)$ & \\
lime & $0.200 (0.40)$ & $0.790 (0.41)$ & $0.936 (0.18)$ & $0.105 (0.19)$ & $0.324 (0.14)$ & $0.041 (0.05)$ & \\
shap & $0.180 (0.38)$ & $0.730 (0.44)$ & $0.966 (0.12)$ & $0.143 (0.23)$ & $0.177 (0.15)$ & $0.041 (0.05)$ & \\
anchor & $0.240 (0.43)$ & $0.560 (0.50)$ & $0.837 (0.31)$ & $0.477 (0.48)$ & $9.167 (47.26)$ & $0.028 (0.03)$ & \\
\end{tabular}
\end{table}

\begin{table}[ht]
\caption{\label{exp-tab:yelp_tree_5}
Comparison on decision tree for Yelp reviews ($p=0.5$, $\epsilon=0.15$).}
\centering
\setlength{\tabcolsep}{2pt}
\begin{tabular}{@{}l|rrrrrrr@{}}
\textbf{} & \texttt{suffic.} $\downarrow$ & \texttt{compreh.} $\uparrow$ & \texttt{robust}. $\uparrow$ & \texttt{aucmorf} $\downarrow$ & \texttt{time (s)} $\downarrow$ & \texttt{proport.} $\downarrow$ & \\
\hline
fred & $0.200 (0.40)$ & $0.560 (0.50)$ & $0.811 (0.28)$ & $0.102 (0.16)$ & $0.357 (0.21)$ & $0.030 (0.03)$ & \\
fredpos & $0.190 (0.39)$ & $0.570 (0.50)$ & $0.818 (0.29)$ & $0.122 (0.19)$ & $0.445 (0.27)$ & $0.030 (0.03)$ & \\
lime & $0.200 (0.40)$ & $0.580 (0.49)$ & $0.830 (0.26)$ & $0.105 (0.19)$ & $0.328 (0.14)$ & $0.030 (0.03)$ & \\
shap & $0.180 (0.38)$ & $0.550 (0.50)$ & $0.972 (0.11)$ & $0.143 (0.23)$ & $0.175 (0.15)$ & $0.030 (0.03)$ & \\
anchor & $0.250 (0.43)$ & $0.560 (0.50)$ & $0.840 (0.31)$ & $0.477 (0.48)$ & $9.164 (47.25)$ & $0.028 (0.03)$ & \\
\end{tabular}
\end{table}

\begin{table}[ht]
\caption{\label{exp-tab:yelp_forest_1}
Comparison on random forest classifier for Yelp reviews ($p=0.1$, $\epsilon=0.15$).}
\centering
\setlength{\tabcolsep}{2pt}
\begin{tabular}{@{}l|rrrrrrr@{}}
\textbf{} & \texttt{suffic.} $\downarrow$ & \texttt{compreh.} $\uparrow$ & \texttt{robust}. $\uparrow$ & \texttt{aucmorf} $\downarrow$ & \texttt{time (s)} $\downarrow$ & \texttt{proport.} $\downarrow$ & \\
\hline
fred & $-0.126 (0.11)$ & $0.107 (0.04)$ & $0.653 (0.28)$ & $0.749 (0.10)$ & $0.306 (0.13)$ & $0.081 (0.08)$ & \\
fredpos & $-0.107 (0.12)$ & $0.100 (0.04)$ & $0.634 (0.30)$ & $0.753 (0.09)$ & $0.328 (0.14)$ & $0.076 (0.09)$ & \\
lime & $-0.138 (0.10)$ & $0.102 (0.04)$ & $0.909 (0.17)$ & $0.782 (0.08)$ & $0.345 (0.13)$ & $0.074 (0.08)$ & \\
shap & $-0.159 (0.09)$ & $0.086 (0.05)$ & $0.968 (0.09)$ & $0.776 (0.08)$ & $0.640 (0.25)$ & $0.076 (0.09)$ & \\
anchor & $-0.050 (0.16)$ & $0.032 (0.05)$ & $0.757 (0.33)$ & $0.942 (0.09)$ & $2.522 (6.23)$ & $0.040 (0.04)$ & \\
\end{tabular}
\end{table}

\begin{table}[ht]
\caption{\label{exp-tab:yelp_distilbert_1}
Comparison on DistilBERT for Yelp reviews ($p=0.1$, $\epsilon=0.15$).}
\centering
\setlength{\tabcolsep}{2pt}
\begin{tabular}{@{}l|rrrrrrr@{}}
\textbf{} & \texttt{suffic.} $\downarrow$ & \texttt{compreh.} $\uparrow$ & \texttt{robust}. $\uparrow$ & \texttt{aucmorf} $\downarrow$ & \texttt{time (s)} $\downarrow$ & \texttt{proport.} $\downarrow$ & \\
\hline
fred & $-0.011 (0.01)$ & $0.012 (0.01)$ & $0.477 (0.19)$ & $0.983 (0.02)$ & $11.324 (3.49)$ & $0.141 (0.12)$ & \\
fredpos & $-0.011 (0.01)$ & $0.007 (0.01)$ & $0.359 (0.16)$ & $0.986 (0.01)$ & $11.816 (3.69)$ & $0.146 (0.13)$ & \\
lime & $-0.012 (0.01)$ & $0.008 (0.01)$ & $0.837 (0.21)$ & $0.986 (0.01)$ & $16.691 (6.84)$ & $0.125 (0.14)$ & \\
shap & $-0.011 (0.01)$ & $0.005 (0.01)$ & $1.000 (0.00)$ & $0.994 (0.01)$ & $2.827 (1.16)$ & $0.146 (0.13)$ & \\
anchor & $-0.007 (0.01)$ & $0.000 (0.00)$ & $0.865 (0.32)$ & $0.999 (0.01)$ & $1.452 (1.11)$ & $0.038 (0.04)$ & \\
\end{tabular}
\end{table}

\begin{table}[ht]
\caption{\label{exp-tab:yelp_distilbert_5}
Comparison on DistilBERT for Yelp reviews ($p=0.5$, $n=70$, $\epsilon=0.15$).}
\centering
\setlength{\tabcolsep}{2pt}
\begin{tabular}{@{}l|rrrrrrr@{}}
\textbf{} & \texttt{suffic.} $\downarrow$ & \texttt{compreh.} $\uparrow$ & \texttt{robust}. $\uparrow$ & \texttt{aucmorf} $\downarrow$ & \texttt{time (s)} $\downarrow$ & \texttt{proport.} $\downarrow$ & \\
\hline
fred & $0.003 (0.01)$ & $0.019 (0.01)$ & $0.517 (0.13)$ & $0.966 (0.01)$ & $16.013 (4.38)$ & $0.222 (0.14)$ & \\
fredpos & $0.007 (0.01)$ & $0.017 (0.01)$ & $0.425 (0.11)$ & $0.975 (0.02)$ & $16.320 (4.36)$ & $0.232 (0.17)$ & \\
lime & $0.004 (0.01)$ & $0.017 (0.01)$ & $0.799 (0.15)$ & $0.973 (0.01)$ & $25.053 (9.54)$ & $0.197 (0.14)$ & \\
shap & $0.006 (0.01)$ & $0.016 (0.01)$ & $1.000 (0.00)$ & $0.973 (0.01)$ & $4.612 (1.58)$ & $0.232 (0.17)$ & \\
anchor & $0.003 (0.01)$ & $0.011 (0.01)$ & $0.520 (0.29)$ & $0.981 (0.01)$ & $184.020 (228.83)$ & $0.134 (0.13)$ & \\
\end{tabular}
\end{table}

\begin{table}[ht]
\caption{\label{exp-tab:yelp_roberta_1}
Comparison on Roberta for Yelp reviews ($p=0.1$, $\epsilon=0.15$).}
\centering
\setlength{\tabcolsep}{2pt}
\begin{tabular}{@{}l|rrrrrrr@{}}
\textbf{} & \texttt{suffic.} $\downarrow$ & \texttt{compreh.} $\uparrow$ & \texttt{robust}. $\uparrow$ & \texttt{aucmorf} $\downarrow$ & \texttt{time (s)} $\downarrow$ & \texttt{proport.} $\downarrow$ & \\
\hline
fred & $0.145 (0.33)$ & $0.180 (0.38)$ & $0.419 (0.29)$ & $0.656 (0.36)$ & $44.358 (7.37)$ & $0.096 (0.09)$ & \\
fredpos & $0.170 (0.34)$ & $0.100 (0.30)$ & $0.393 (0.38)$ & $0.716 (0.33)$ & $45.003 (8.14)$ & $0.065 (0.07)$ & \\
lime & $0.103 (0.28)$ & $0.100 (0.30)$ & $0.601 (0.34)$ & $0.676 (0.39)$ & $69.072 (21.34)$ & $0.065 (0.07)$ & \\
shap & $0.244 (0.40)$ & $0.031 (0.17)$ & $1.000 (0.00)$ & $0.770 (0.31)$ & $8.455 (3.40)$ & $0.065 (0.07)$ & \\
anchor & $0.419 (0.43)$ & $0.050 (0.22)$ & $0.626 (0.43)$ & $0.926 (0.25)$ & $22.924 (47.58)$ & $0.033 (0.04)$ & \\
\end{tabular}
\end{table}

\begin{table}[ht]
\caption{\label{exp-tab:imdb_logistic_1}
Comparison on logistic classifier for IMDb ($p=0.1$, $\epsilon=0.15$).}
\centering
\setlength{\tabcolsep}{2pt}
\begin{tabular}{@{}l|rrrrrrr@{}}
\textbf{} & \texttt{suffic.} $\downarrow$ & \texttt{compreh.} $\uparrow$ & \texttt{robust}. $\uparrow$ & \texttt{aucmorf} $\downarrow$ & \texttt{time (s)} $\downarrow$ & \texttt{proport.} $\downarrow$ & \\
\hline
fred & $-0.122 (0.05)$ & $0.045 (0.02)$ & $0.465 (0.11)$ & $0.873 (0.05)$ & $0.194 (0.04)$ & $0.151 (0.08)$ & \\
fredpos & $-0.117 (0.05)$ & $0.043 (0.02)$ & $0.416 (0.09)$ & $0.872 (0.05)$ & $0.211 (0.04)$ & $0.146 (0.09)$ & \\
lime & $-0.141 (0.05)$ & $0.048 (0.02)$ & $0.947 (0.10)$ & $0.908 (0.03)$ & $0.204 (0.03)$ & $0.142 (0.09)$ & \\
shap & $-0.141 (0.05)$ & $0.047 (0.02)$ & $1.000 (0.00)$ & $0.876 (0.05)$ & $0.202 (0.30)$ & $0.146 (0.09)$ & \\
anchor & $-0.090 (0.09)$ & $0.020 (0.02)$ & $0.591 (0.35)$ & $0.964 (0.03)$ & $0.671 (0.74)$ & $0.056 (0.04)$ & \\
\end{tabular}
\end{table}

\begin{table}[ht]
\caption{\label{exp-tab:imdb_logistic_5}
Comparison on logistic classifier for IMDb ($p=0.5$, $\epsilon=0.15$).}
\centering
\setlength{\tabcolsep}{2pt}
\begin{tabular}{@{}l|rrrrrrr@{}}
\textbf{} & \texttt{suffic.} $\downarrow$ & \texttt{compreh.} $\uparrow$ & \texttt{robust}. $\uparrow$ & \texttt{aucmorf} $\downarrow$ & \texttt{time (s)} $\downarrow$ & \texttt{proport.} $\downarrow$ & \\
\hline
fred & $-0.103 (0.04)$ & $0.066 (0.02)$ & $0.595 (0.13)$ & $0.876 (0.05)$ & $0.603 (0.12)$ & $0.294 (0.13)$ & \\
fredpos & $-0.097 (0.04)$ & $0.063 (0.02)$ & $0.539 (0.10)$ & $0.874 (0.05)$ & $0.687 (0.14)$ & $0.296 (0.12)$ & \\
lime & $-0.138 (0.05)$ & $0.052 (0.02)$ & $0.950 (0.10)$ & $0.911 (0.03)$ & $0.375 (0.05)$ & $0.175 (0.10)$ & \\
shap & $-0.112 (0.04)$ & $0.066 (0.02)$ & $1.000 (0.00)$ & $0.880 (0.04)$ & $0.336 (0.44)$ & $0.296 (0.12)$ & \\
anchor & $-0.099 (0.08)$ & $0.020 (0.02)$ & $0.622 (0.31)$ & $0.964 (0.03)$ & $1.213 (1.53)$ & $0.057 (0.04)$ & \\
\end{tabular}
\end{table}

\begin{table}[ht]
\caption{\label{exp-tab:imdb_tree_1}
Comparison on decision tree for IMDb ($p=0.1$, $\epsilon=0.15$).}
\centering
\setlength{\tabcolsep}{2pt}
\begin{tabular}{@{}l|rrrrrrr@{}}
\textbf{} & \texttt{suffic.} $\downarrow$ & \texttt{compreh.} $\uparrow$ & \texttt{robust}. $\uparrow$ & \texttt{aucmorf} $\downarrow$ & \texttt{time (s)} $\downarrow$ & \texttt{proport.} $\downarrow$ & \\
\hline
fred & $0.250 (0.43)$ & $0.690 (0.46)$ & $0.792 (0.29)$ & $0.174 (0.24)$ & $0.145 (0.03)$ & $0.059 (0.04)$ & \\
fredpos & $0.270 (0.44)$ & $0.530 (0.50)$ & $0.749 (0.35)$ & $0.341 (0.36)$ & $0.160 (0.03)$ & $0.048 (0.03)$ & \\
lime & $0.200 (0.40)$ & $0.550 (0.50)$ & $0.820 (0.27)$ & $0.091 (0.08)$ & $0.199 (0.02)$ & $0.048 (0.03)$ & \\
shap & $0.160 (0.37)$ & $0.480 (0.50)$ & $0.933 (0.19)$ & $0.146 (0.19)$ & $0.185 (0.13)$ & $0.048 (0.03)$ & \\
anchor & $0.150 (0.36)$ & $0.460 (0.50)$ & $0.777 (0.30)$ & $0.542 (0.48)$ & $1.946 (8.13)$ & $0.035 (0.01)$ & \\
\end{tabular}
\end{table}

\begin{table}[ht]
\caption{\label{exp-tab:imdb_tree_5}
Comparison on decision tree for IMDb ($p=0.5$, $\epsilon=0.15$).}
\centering
\setlength{\tabcolsep}{2pt}
\begin{tabular}{@{}l|rrrrrrr@{}}
\textbf{} & \texttt{suffic.} $\downarrow$ & \texttt{compreh.} $\uparrow$ & \texttt{robust}. $\uparrow$ & \texttt{aucmorf} $\downarrow$ & \texttt{time (s)} $\downarrow$ & \texttt{proport.} $\downarrow$ & \\
\hline
fred & $0.160 (0.37)$ & $0.440 (0.50)$ & $0.736 (0.30)$ & $0.098 (0.09)$ & $0.172 (0.03)$ & $0.039 (0.02)$ & \\
fredpos & $0.220 (0.41)$ & $0.420 (0.49)$ & $0.855 (0.26)$ & $0.304 (0.34)$ & $0.213 (0.04)$ & $0.035 (0.02)$ & \\
lime & $0.190 (0.39)$ & $0.470 (0.50)$ & $0.726 (0.31)$ & $0.091 (0.08)$ & $0.200 (0.02)$ & $0.035 (0.02)$ & \\
shap & $0.180 (0.38)$ & $0.440 (0.50)$ & $0.921 (0.22)$ & $0.146 (0.19)$ & $0.182 (0.14)$ & $0.035 (0.02)$ & \\
anchor & $0.180 (0.38)$ & $0.460 (0.50)$ & $0.775 (0.31)$ & $0.542 (0.48)$ & $1.972 (8.24)$ & $0.033 (0.01)$ & \\
\end{tabular}
\end{table}

\begin{table}[ht]
\caption{\label{exp-tab:imdb_forest_1}
Comparison on random forest classifier for IMDb ($p=0.1$, $\epsilon=0.15$).}
\centering
\setlength{\tabcolsep}{2pt}
\begin{tabular}{@{}l|rrrrrrr@{}}
\textbf{} & \texttt{suffic.} $\downarrow$ & \texttt{compreh.} $\uparrow$ & \texttt{robust}. $\uparrow$ & \texttt{aucmorf} $\downarrow$ & \texttt{time (s)} $\downarrow$ & \texttt{proport.} $\downarrow$ & \\
\hline
fred & $-0.141 (0.09)$ & $0.091 (0.03)$ & $0.489 (0.18)$ & $0.840 (0.09)$ & $0.319 (0.03)$ & $0.139 (0.10)$ & \\
fredpos & $-0.136 (0.09)$ & $0.069 (0.03)$ & $0.401 (0.23)$ & $0.847 (0.08)$ & $0.324 (0.03)$ & $0.122 (0.10)$ & \\
lime & $-0.163 (0.09)$ & $0.082 (0.03)$ & $0.904 (0.16)$ & $0.861 (0.06)$ & $0.363 (0.02)$ & $0.111 (0.07)$ & \\
shap & $-0.173 (0.10)$ & $0.072 (0.03)$ & $0.961 (0.11)$ & $0.821 (0.08)$ & $0.822 (0.22)$ & $0.122 (0.10)$ & \\
anchor & $-0.095 (0.11)$ & $0.007 (0.03)$ & $0.813 (0.32)$ & $0.987 (0.06)$ & $0.995 (2.98)$ & $0.043 (0.03)$ & \\
\end{tabular}
\end{table}

\begin{table}[ht]
\caption{\label{exp-tab:imdb_forest_5}
Comparison on random forest classifier for IMDb ($p=0.5$, $\epsilon=0.15$).}
\centering
\setlength{\tabcolsep}{2pt}
\begin{tabular}{@{}l|rrrrrrr@{}}
\textbf{} & \texttt{suffic.} $\downarrow$ & \texttt{compreh.} $\uparrow$ & \texttt{robust}. $\uparrow$ & \texttt{aucmorf} $\downarrow$ & \texttt{time (s)} $\downarrow$ & \texttt{proport.} $\downarrow$ & \\
\hline
fred & $-0.175 (0.09)$ & $0.124 (0.04)$ & $0.582 (0.17)$ & $0.904 (0.11)$ & $0.780 (0.14)$ & $0.299 (0.14)$ & \\
fredpos & $-0.141 (0.10)$ & $0.077 (0.05)$ & $0.639 (0.25)$ & $0.870 (0.10)$ & $0.736 (0.16)$ & $0.215 (0.19)$ & \\
lime & $-0.158 (0.11)$ & $0.076 (0.04)$ & $0.916 (0.15)$ & $0.861 (0.06)$ & $0.589 (0.04)$ & $0.113 (0.09)$ & \\
shap & $-0.167 (0.11)$ & $0.086 (0.05)$ & $0.972 (0.06)$ & $0.821 (0.08)$ & $1.390 (0.35)$ & $0.215 (0.19)$ & \\
anchor & $-0.093 (0.11)$ & $0.008 (0.03)$ & $0.807 (0.32)$ & $0.987 (0.06)$ & $1.719 (5.10)$ & $0.046 (0.04)$ & \\
\end{tabular}
\end{table}

\begin{table}[ht]
\caption{\label{exp-tab:imdb_distilbert_1}
Comparison on DistilBERT for IMDb ($p=0.1$, $\epsilon=0.15$).}
\centering
\setlength{\tabcolsep}{2pt}
\begin{tabular}{@{}l|rrrrrrr@{}}
\textbf{} & \texttt{suffic.} $\downarrow$ & \texttt{compreh.} $\uparrow$ & \texttt{robust}. $\uparrow$ & \texttt{aucmorf} $\downarrow$ & \texttt{time (s)} $\downarrow$ & \texttt{proport.} $\downarrow$ & \\
\hline
fred & $-0.011 (0.01)$ & $0.007 (0.00)$ & $0.460 (0.18)$ & $0.993 (0.01)$ & $9.092 (0.73)$ & $0.155 (0.09)$ & \\
fredpos & $-0.011 (0.01)$ & $0.004 (0.00)$ & $0.344 (0.15)$ & $0.990 (0.01)$ & $9.346 (0.92)$ & $0.167 (0.08)$ & \\
lime & $-0.014 (0.01)$ & $0.005 (0.00)$ & $0.872 (0.18)$ & $0.991 (0.01)$ & $11.029 (1.29)$ & $0.119 (0.07)$ & \\
shap & $-0.012 (0.01)$ & $0.002 (0.01)$ & $1.000 (0.00)$ & $0.995 (0.01)$ & $1.827 (0.56)$ & $0.167 (0.08)$ & \\
anchor & $-0.008 (0.01)$ & $-0.001 (0.00)$ & $0.990 (0.10)$ & $1.001 (0.01)$ & $0.645 (0.09)$ & $0.044 (0.03)$ & \\
\end{tabular}
\end{table}

\begin{table}[ht]
\caption{\label{exp-tab:imdb_roberta_1}
Comparison on Roberta for IMDb ($p=0.1$, $\epsilon=0.15$).}
\centering
\setlength{\tabcolsep}{2pt}
\begin{tabular}{@{}l|rrrrrrr@{}}
\textbf{} & \texttt{suffic.} $\downarrow$ & \texttt{compreh.} $\uparrow$ & \texttt{robust}. $\uparrow$ & \texttt{aucmorf} $\downarrow$ & \texttt{time (s)} $\downarrow$ & \texttt{proport.} $\downarrow$ & \\
\hline
fred & $0.278 (0.44)$ & $0.488 (0.49)$ & $0.635 (0.35)$ & $0.353 (0.36)$ & $45.841 (3.75)$ & $0.090 (0.06)$ & \\
fredpos & $0.417 (0.48)$ & $0.266 (0.44)$ & $0.674 (0.35)$ & $0.454 (0.35)$ & $46.691 (4.01)$ & $0.060 (0.03)$ & \\
lime & $0.210 (0.40)$ & $0.278 (0.44)$ & $0.855 (0.28)$ & $0.295 (0.33)$ & $65.123 (7.64)$ & $0.060 (0.03)$ & \\
shap & $0.436 (0.48)$ & $0.159 (0.36)$ & $1.000 (0.00)$ & $0.475 (0.32)$ & $15.901 (5.05)$ & $0.060 (0.03)$ & \\
anchor & $0.453 (0.48)$ & $0.228 (0.42)$ & $0.642 (0.41)$ & $0.777 (0.40)$ & $54.867 (112.66)$ & $0.038 (0.02)$ & \\
\end{tabular}
\end{table}

\begin{table}[t]
\caption{\label{exp-tab:tweets_tree_1}
Comparison on decision tree for tweets hate speech detection ($p=0.1$, $\epsilon=0.15$).}
\centering
\setlength{\tabcolsep}{2pt}
\begin{tabular}{@{}l|rrrrrrr@{}}
\textbf{} & \texttt{suffic.} $\downarrow$ & \texttt{compreh.} $\uparrow$ & \texttt{robust}. $\uparrow$ & \texttt{aucmorf} $\downarrow$ & \texttt{time (s)} $\downarrow$ & \texttt{proport.} $\downarrow$ & \\
\hline
fred & $0.870 (0.34)$ & $0.950 (0.22)$ & $0.982 (0.10)$ & $0.078 (0.04)$ & $0.046 (0.01)$ & $0.143 (0.05)$ & \\
fredpos & $0.880 (0.32)$ & $1.000 (0.00)$ & $0.952 (0.15)$ & $0.078 (0.04)$ & $0.052 (0.01)$ & $0.148 (0.05)$ & \\
lime & $0.870 (0.34)$ & $0.990 (0.10)$ & $0.907 (0.21)$ & $0.082 (0.06)$ & $0.125 (0.01)$ & $0.148 (0.05)$ & \\
shap & $0.880 (0.32)$ & $0.970 (0.17)$ & $1.000 (0.00)$ & $0.088 (0.08)$ & $0.035 (0.18)$ & $0.148 (0.05)$ & \\
anchor & $0.880 (0.32)$ & $0.850 (0.36)$ & $0.669 (0.39)$ & $0.109 (0.14)$ & $0.578 (0.47)$ & $0.144 (0.05)$ & \\
\end{tabular}
\end{table}

\begin{table}[ht]
\caption{\label{exp-tab:tweets_forest_1}
Comparison on random forest classifier for tweets hate speech detection ($p=0.1$, $\epsilon=0.15$).}
\centering
\setlength{\tabcolsep}{2pt}
\begin{tabular}{@{}l|rrrrrrr@{}}
\textbf{} & \texttt{suffic.} $\downarrow$ & \texttt{compreh.} $\uparrow$ & \texttt{robust}. $\uparrow$ & \texttt{aucmorf} $\downarrow$ & \texttt{time (s)} $\downarrow$ & \texttt{proport.} $\downarrow$ & \\
\hline
fred & $0.782 (0.21)$ & $0.391 (0.15)$ & $0.976 (0.09)$ & $0.162 (0.06)$ & $0.273 (0.06)$ & $0.108 (0.03)$ & \\
fredpos & $0.784 (0.20)$ & $0.388 (0.15)$ & $0.985 (0.09)$ & $0.166 (0.06)$ & $0.250 (0.07)$ & $0.109 (0.03)$ & \\
lime & $0.784 (0.20)$ & $0.385 (0.14)$ & $0.974 (0.14)$ & $0.159 (0.05)$ & $0.453 (0.05)$ & $0.109 (0.03)$ & \\
shap & $0.784 (0.20)$ & $0.384 (0.15)$ & $1.000 (0.00)$ & $0.156 (0.05)$ & $0.153 (0.14)$ & $0.109 (0.03)$ & \\
anchor & $0.788 (0.20)$ & $0.281 (0.19)$ & $0.493 (0.40)$ & $0.221 (0.10)$ & $7.841 (4.00)$ & $0.109 (0.03)$ & \\
\end{tabular}
\end{table}

\begin{table}[ht]
\caption{\label{exp-tab:tweets_distilbert_1}
Comparison on DistilBERT for tweets hate speech detection ($p=0.1$, $\epsilon=0.15$).}
\centering
\setlength{\tabcolsep}{2pt}
\begin{tabular}{@{}l|rrrrrrr@{}}
\textbf{} & \texttt{suffic.} $\downarrow$ & \texttt{compreh.} $\uparrow$ & \texttt{robust}. $\uparrow$ & \texttt{aucmorf} $\downarrow$ & \texttt{time (s)} $\downarrow$ & \texttt{proport.} $\downarrow$ & \\
\hline
fred & $-0.002 (0.01)$ & $0.019 (0.01)$ & $0.691 (0.21)$ & $0.974 (0.01)$ & $8.309 (0.88)$ & $0.411 (0.10)$ & \\
fredpos & $0.001 (0.01)$ & $0.012 (0.01)$ & $0.460 (0.16)$ & $0.978 (0.01)$ & $8.392 (0.91)$ & $0.450 (0.13)$ & \\
lime & $-0.005 (0.01)$ & $0.016 (0.01)$ & $0.979 (0.07)$ & $0.972 (0.01)$ & $9.414 (1.11)$ & $0.427 (0.10)$ & \\
shap & $-0.003 (0.01)$ & $0.013 (0.01)$ & $1.000 (0.00)$ & $0.978 (0.01)$ & $0.497 (0.33)$ & $0.450 (0.13)$ & \\
anchor & $-0.002 (0.01)$ & $0.012 (0.01)$ & $0.891 (0.20)$ & $0.978 (0.01)$ & $11.732 (11.18)$ & $0.307 (0.17)$ & \\
\end{tabular}
\end{table}

\end{document}